\documentclass[sigconf]{acmart}
\AtBeginDocument{%
  }

\copyrightyear{2026}
\acmYear{2026}
\setcopyright{cc}
\setcctype{by}
\acmConference[MM '26] {Proceedings of the 34th ACM International Conference on Multimedia}{November 10--14, 2026}{Rio de Janeiro, Brazil.}
\acmBooktitle{Proceedings of the 34th ACM International Conference on Multimedia (MM '26), November 10--14, 2026, Rio de Janeiro, Brazil}
\acmISBN{979-8-4007-2213-4/2026/11}
\acmDOI{10.1145/3767308.3835364}

\acmSubmissionID{mfp2554}

\usepackage{multirow} 
\usepackage{amsmath} 
\usepackage{graphicx} 
\usepackage{bbding} 
\usepackage{wrapfig} 
\usepackage{colortbl} 
\usepackage{caption} 
\usepackage{array} 
\usepackage{diagbox} 
\usepackage{booktabs}       
\usepackage{amsfonts}       
\usepackage{microtype}      
\usepackage{tabularx} 
\usepackage{pifont} 
\usepackage{cuted}
\usepackage{subcaption}
\usepackage{tikz}
\usepackage{makecell}
\usepackage{balance} 
\begin{document}

\title{Controllable Video Object Insertion via Multi-View Priors}


\author{
Qi Xia
}
\authornote{Equal contribution.}
\orcid{0009-0005-2029-7870}
\affiliation{%
  \institution{ShanghaiTech University}
  \city{Shanghai}
  \country{China}
}
\email{xiaqi2025@shanghaitech.edu.cn}

\author{
Peishan Cong
}
\authornotemark[1]
\orcid{0000-0001-8853-387X}
\affiliation{%
  \institution{ShanghaiTech University}
  \city{Shanghai}
  \country{China}
}
\email{congpsh@shanghaitech.edu.cn}

\author{
Yichen Yao
}
\orcid{0009-0006-9319-9339}
\affiliation{%
  \institution{ShanghaiTech University}
  \city{Shanghai}
  \country{China}
  }
\email{yaoych2023@shanghaitech.edu.cn}

\author{
Ziyi Wang
}
\orcid{0009-0002-0841-3929}
\affiliation{%
  \institution{ShanghaiTech University}
  \city{Shanghai}
  \country{China}
  }
\email{wangzy17@shanghaitech.edu.cn}

\author{
Yaoqin Ye
}
\orcid{0009-0006-3989-4170}
\affiliation{%
  \institution{ShanghaiTech University}
  \city{Shanghai}
  \country{China}
}
\email{yeyq2025@shanghaitech.edu.cn}

\author{
Yuexin Ma
}
\orcid{0000-0001-7237-988X}
\authornote{Corresponding author.}
\affiliation{%
  \institution{ShanghaiTech University}
  \city{Shanghai}
  \country{China}
  }
\email{mayuexin@shanghaitech.edu.cn}

\renewcommand{\shortauthors}{Xia et al.}

\begin{abstract}

Video object insertion places a user-specified object in an existing dynamic scene. Existing methods typically condition generation on text or a single reference image. Consequently, object appearance is underconstrained under viewpoint changes, often leading to identity drift, incorrect foreground-background layering, boundary artifacts, and temporal flickering. In this paper, we propose a video object insertion framework that incorporates multi-view object priors to address these limitations. The framework lifts a 2D reference image into a multi-view representation and uses view-consistent conditioning to provide stable identity guidance and view-adaptive appearance cues. A quality-aware weighting mechanism reduces the influence of noisy or imperfect reconstructed views. We further introduce an Integration-Aware Consistency Module that promotes plausible occlusion, clean boundaries, and temporal continuity. Experiments demonstrate that the proposed framework improves visual quality, controllability, identity consistency, and foreground-background integration for video object insertion compared to the baseline methods. \textit{Project page: }\url{https://polarisxq.github.io/MOVI/}.

\end{abstract}



\begin{teaserfigure}
  \centering
  \vspace{-3ex}
  \includegraphics[width=\linewidth]{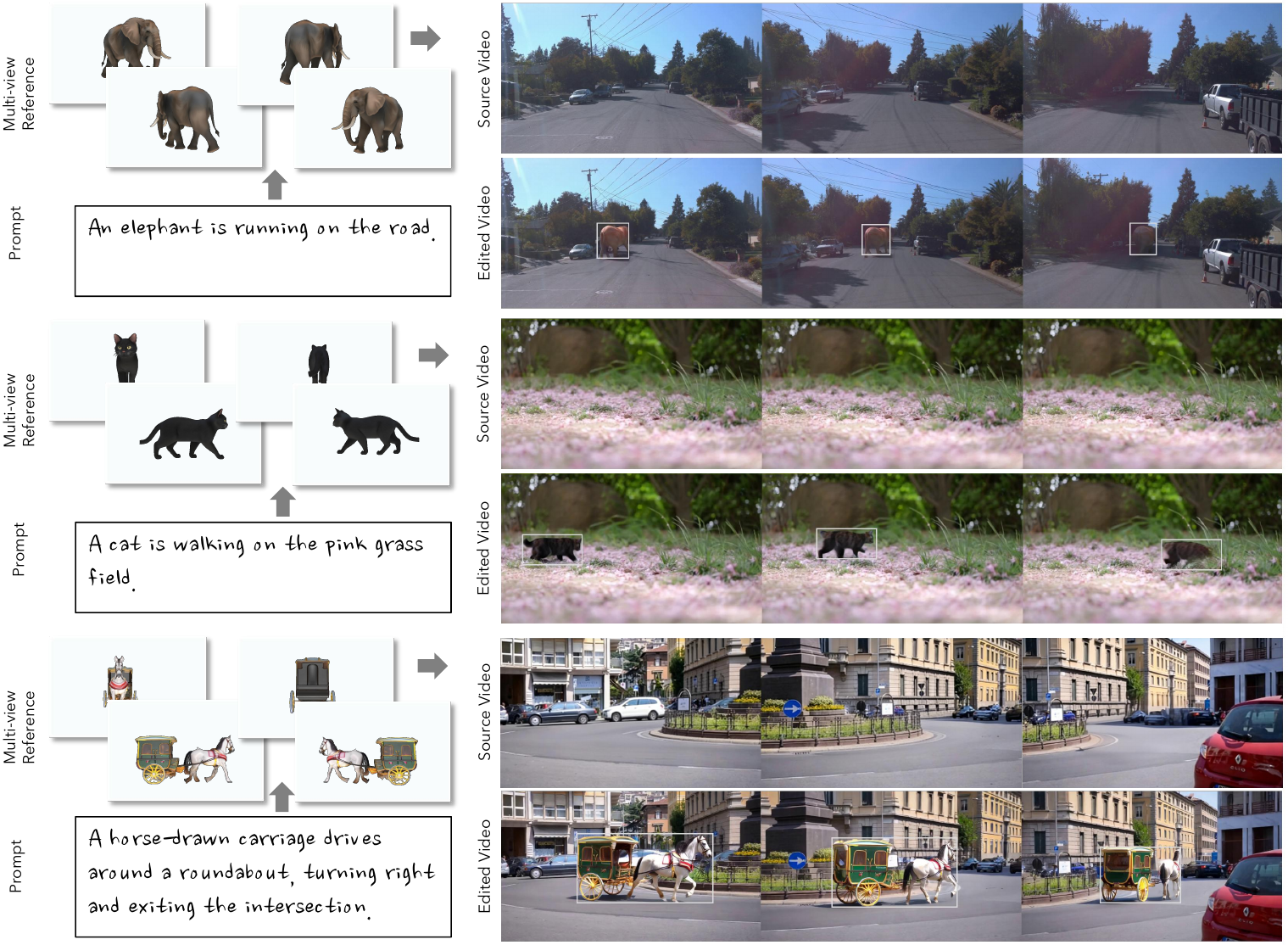}
  \vspace{-6ex}
  \caption{The proposed controllable video object insertion framework uses 3D-derived multi-view priors, view-consistent conditioning module, and integration-aware consistency module to preserve inserted object's identity and improve foreground-background integration in dynamic scenes. Source/background frames: top, \textcopyright\ Waymo LLC, Waymo Open Dataset~\cite{waymo_open_dataset}; middle, \textcopyright\ Ke Aierxi, used with permission; bottom, \textcopyright\ 2017 DAVIS authors, CC BY-NC 4.0~\cite{perazzi2016davis}.}
  \label{fig:teaser}
\end{teaserfigure}

\maketitle

\section{Introduction}
Video object insertion integrates new assets into existing dynamic environments, enabling realistic and controllable video editing. It is relevant to applications such as film post-production, immersive AR/VR content creation, and autonomous driving simulation~\cite{Xing2023ASO, gao2025gallery, bai2024anything, yang2023unisim}, where inserted objects must remain visually coherent with complex scenes over time.

Early video generation methods~\cite{Hong2022CogVideoLP, Blattmann2023AlignYL, Gupta2023PhotorealisticVG, Yang2024CogVideoXTD, HaCohen2024LTXVideo, Kong2024HunyuanVideoAS, wan2025} primarily synthesize complete videos from text prompts and improve global realism and motion quality. However, they are not designed for object insertion, which requires both precise foreground control and compatibility with the original background. A successful insertion must provide accurate spatial alignment, plausible occlusion, and temporally consistent appearance.

Recent works~\cite{MagicMotion, vace, gao2025loraedit, yang2025mtv, yatim2025dynvfx, hu2025hunyuancustommultimodaldrivenarchitecturecustomized,yang2026gencompositor} improve controllability by introducing spatial conditions, including point trajectories~\cite{yang2026gencompositor}, bounding boxes~\cite{MagicMotion, vace, gao2025loraedit,yang2025mtv}, and masks~\cite{MagicMotion, hu2025hunyuancustommultimodaldrivenarchitecturecustomized, yatim2025dynvfx}. Nevertheless, these methods typically use text~\cite{yatim2025dynvfx} or a single reference image~\cite{MagicMotion, vace, gao2025loraedit,yang2025mtv,hu2025hunyuancustommultimodaldrivenarchitecturecustomized} to specify object appearance. Such partial observations can cause \textbf{identity drift, geometric distortion, and artifacts at unseen viewpoints}. In addition, many methods do not model the foreground-background interaction. Camera motion and viewpoint changes can therefore lead to \textbf{incorrect layering, boundary artifacts, and temporal flickering}. The central challenge is to preserve the inserted object's identity across different viewpoints while achieving perceptually plausible integration into dynamic scenes.
In this paper, we propose a video object insertion framework that combines multi-view object priors with integration-aware consistency to address appearance inconsistency and integration artifacts in dynamic scenes. Specifically, we construct a multi-view representation by lifting the object into 3D and rendering it from multiple viewpoints, thereby providing the generative model with broader appearance information. To exploit this information, we design a \textbf{View-Consistent Conditioning Module}. First, \textit{Identity-Preserving Latent Injection} fuses multi-view frames with background latents to provide stable identity guidance. Second, a \textit{Multi-View Feature Bank} serves as external memory for object details. Through cross-attention, the model retrieves viewpoint-relevant textures and geometric cues from this feature bank.
To improve robustness, we implement a \textit{Quality-Aware Weighting} mechanism. During training, we retain naturally imperfect renderings from the 3D lifting pipeline, including views with missing geometry, over-smoothed surfaces, and warped textures. Cross-modal consistency scores estimate the reliability of each view and adaptively reweight its contribution, suppressing unreliable renderings while retaining useful multi-view evidence. We further design an \textbf{Integration-Aware Consistency Module} to improve spatial realism. A Geometric Grounding module includes a Depth Head for depth-aware occlusion reasoning and a Contour Head for sharper object boundaries. Temporal Consistency Optimization reduces temporal flickering across frames. Together, these components promote stable and realistic video object insertion.

We evaluate the framework on multiple datasets using reference-based metrics, reference-free perceptual metrics, targeted ablations, and a user study. The results demonstrate the effectiveness of the individual components and show that the framework improves identity preservation, controllability, temporal quality, and perceptual foreground-background integration over the baseline methods.

Our primary contributions are summarized as follows: 

\begin{itemize}
    \item We present a video object insertion framework that uses multi-view priors to improve appearance consistency and the realism of foreground-background integration while supporting both coarse bounding-box control and fine-grained mask control.
    
    \item We propose a \textbf{View-Consistent Conditioning Module}, which enables the model to retrieve and integrate viewpoint-relevant, reliable object features from multi-view priors for view-consistent generation.
    
    \item We introduce an \textbf{Integration-Aware Consistency Module} with auxiliary depth and contour heads and a temporal constraint to improve spatial layering, object boundaries, and inter-frame stability.
    
    \item Extensive experiments on multiple benchmarks demonstrate improved visual quality, controllability, and multi-view consistency relative to existing methods.
\end{itemize}

\section{Related Works}

\subsection{Generative Video Editing}
With recent advances in video diffusion models, generative video editing has emerged as a powerful approach to manipulating existing video content.
Early, primarily text-driven methods focused on global style transfer and full-frame semantic editing~\cite{Wu_2023_Tune-A-Video, geyer2023tokenflow, li2024vidtome, wu2024fairy, song2024save, liu2024videop2p, wang2023zero, cong2023flatten, yang2023rerender}. Although these methods can substantially alter the visual appearance or semantics of a video, they do not provide the fine-grained control required for object-level editing.
Later work introduced first-frame-guided editing, in which an edited first frame is propagated through the video. AnyV2V~\cite{ku_anyv2v} and I2VEdit~\cite{ouyang2024i2vedit} inject spatial and temporal features into the diffusion process to reconstruct motion and refine object appearance. UniVideo~\cite{wei2025univideo} further introduces fine-grained conditioning signals to promote temporal coherence. These methods offer effective control of object motion and appearance, but they are primarily designed for subject replacement or attribute editing.
Video inpainting methods~\cite{propainter, xu2019deep, ecfvi, floed, green, cococo, avid} fill missing regions with background-consistent content. They are generally intended for restoration rather than for inserting novel, user-specified foreground objects, and therefore remain limited for object insertion into dynamic scenes. In this work, we focus on inserting novel objects while preserving geometric plausibility and appropriate interaction with the background.

\subsection{Controllable Video Object Insertion}
Video object insertion places a novel object in an original video while maintaining compatibility with the background. Several frameworks~\cite{zhou2024migc,chen2024anydoor,wang2023videocomposer,winter2024objectdrop, vace, yang2025mtv, gao2025loraedit} incorporate spatiotemporal guidance signals to guide object insertion. MagicMotion~\cite{MagicMotion} introduces dense-to-sparse trajectory guidance, which supports masks or bounding-box sequences to specify object trajectories. VACE~\cite{vace} unifies multiple tasks through a Video Condition Unit, supporting masked editing and object insertion. MTV-Inpaint~\cite{yang2025mtv} combines scene completion and object insertion through a dual-branch spatial-attention mechanism. LoRA-Edit~\cite{gao2025loraedit} adapts pre-trained image-to-video models through mask-aware LoRA fine-tuning, using spatiotemporal masks to preserve unedited regions and control the motion and appearance evolution of edited content.

Despite this progress, these methods often rely on a single reference image and cannot fully capture the appearance of an inserted object from different viewpoints. This limitation can lead to geometric distortion and appearance drift under substantial camera motion or viewpoint changes. Existing methods also do not explicitly model depth relationships or occlusion constraints, which can result in implausible foreground-background blending and blurred object boundaries in complex scenes.

\subsection{3D-Aware Video Synthesis}

The absence of explicit 3D geometric understanding in 2D generative video models can lead to viewpoint inconsistency, incorrect perspective, and implausible occlusion. 3D-aware video synthesis addresses these issues by incorporating geometric priors into generative models~\cite{chen2023geodiffusion,bai2024anything,liu2024place,yao2025uni4d,jin2025insertanywhere,gao2026pisco}.

Recent 3D-aware video editing methods use monocular depth estimation as a weak control signal~\cite{gao2026pisco} or combine 4D scene reconstruction with view-consistent editing~\cite{jin2025insertanywhere}. However, they use depth without jointly modeling occlusion and object contours, may require computationally expensive full-scene reconstruction, and lack dedicated mechanisms for injecting multi-view references for object insertion. Consequently, appearance consistency remains challenging under large viewpoint changes.

Recent studies have explored multi-view object references to improve geometric and appearance consistency under viewpoint changes. MVHOI~\cite{tong2026mvhoi} combines a Unified Object Anchor built on a 3D foundation model with multi-view appearance retrieval to support human-object interaction reenactment under complex non-planar object motion. However, it is specialized for human-object interaction reenactment and does not explicitly model the foreground-background depth relationships or boundary integration required for general video object insertion.

In this work, we use multi-view information for the inserted object. Combined with a view-consistent conditioning module and an integration-aware consistency module, the proposed framework supports 3D-aware video object insertion without full-scene reconstruction and accommodates both coarse bounding-box control and fine-grained mask control.

\section{Method}

\subsection{Overview}
Given a source video sequence $\mathcal{V}_{\mathrm{src}}=\{I_{\mathrm{src}}^{t}\}_{t=1}^{T}$, where $I_{\mathrm{src}}^{t}\in\mathbb{R}^{H\times W\times3}$, a foreground reference image $I_{\mathrm{ref}}\in\mathbb{R}^{H\times W\times3}$, a text description $\mathcal{P}$, and spatiotemporal control signals $\mathcal{C}=\{C^{t}\}_{t=1}^{T}$, our objective is to generate a video sequence $\mathcal{V}_{\mathrm{gen}}=\{I_{\mathrm{gen}}^{t}\}_{t=1}^{T}$ that inserts the specified foreground into the designated mask or bounding-box regions. The primary setting uses a user-provided reference image, matching the inputs available to reference-guided baselines. In the practical text-driven setting, a text-to-image model first synthesizes $I_{\mathrm{ref}}$ from $\mathcal{P}$, while the remaining insertion pipeline is unchanged. In both settings, the generated video should preserve the original background and maintain foreground identity, plausible occlusion relationships, clean object boundaries, and temporal coherence.

The framework comprises three modules. The \textbf{Multi-View Reference Generation Module} lifts $I_{\mathrm{ref}}$ into multi-view renderings. The \textbf{View-Consistent Conditioning Module} injects the resulting multi-view information via \textit{Identity-Preserving Latent Injection} and a \textit{Multi-View Feature Bank}. This module further performs a \textit{Semantic Consistency Check} to reduce the influence of unreliable reconstructed views. The \textbf{Integration-Aware Consistency Module} combines latent depth and contour supervision with temporal optimization to improve foreground-background layering, boundary quality, and motion continuity.

\begin{figure*}[t]
    \centering
    \includegraphics[width=\linewidth]{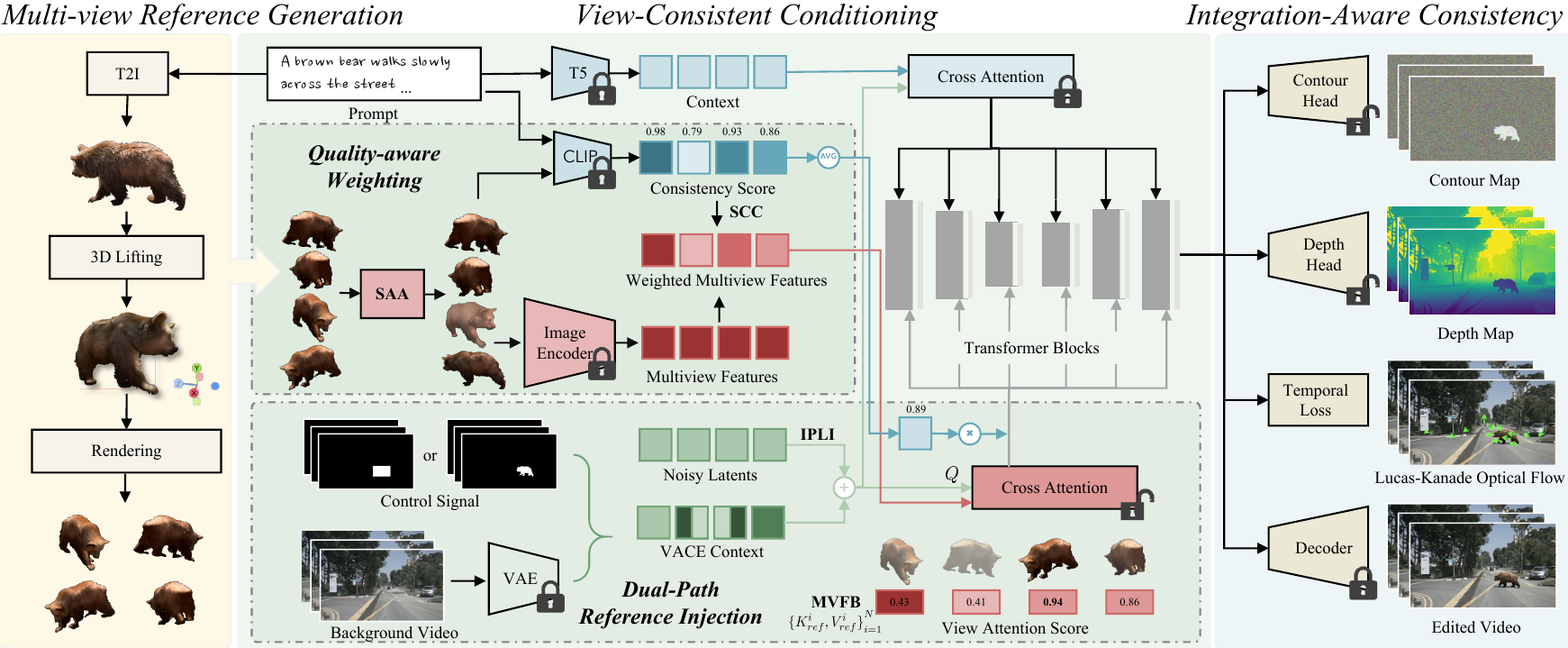}
    \vspace{-5ex}
    \caption{Framework overview. IPLI: Identity-Preserving Latent Injection; MVFB: Multi-View Feature Bank; SAA: Synthesized Artifact Augmentation; SCC: Semantic Consistency Check.}
    \label{fig:framework}
    \vspace{-1ex}
\end{figure*}

\subsection{Multi-view Reference Generation}
The single-view reference image $I_{\mathrm{ref}}$ provides only partial information about the foreground object, its appearance and geometry from unobserved perspectives are not explicitly provided. The absence of multi-view information can lead to inconsistency and distortion when the object is viewed from different angles.

To address this information bottleneck, we use Hunyuan3D 2.0~\cite{hunyuan3d22025tencent} to lift $I_{\mathrm{ref}}$ into a textured mesh $\mathcal{M}$. We then render the multi-view image set $\mathcal{R}_{\mathrm{mv}}=\{I_{\mathrm{mv}}^{n}\}_{n=1}^{N}$ using uniformly distributed camera extrinsics. Hunyuan3D is a preprocessing tool rather than a required architectural component: users may provide multi-view images directly, and alternative reconstruction systems can replace it without changing the conditioning architecture. The resulting views provide the generative model with richer information for object insertion across multiple viewpoints.

\subsection{View-Consistent Conditioning}

To incorporate rendered multi-view priors without compromising generation stability, we introduce the View-Consistent Conditioning Module. It consists of two components: \textit{Dual-Path Reference Injection}, which provides complementary identity-preserving and view-adaptive appearance guidance, and \textit{Quality-Aware Weighting}, which reduces the influence of unreliable 3D renderings.

\subsubsection{Dual-Path Reference Injection}

Simply replacing a single-view reference with stacked multi-view renderings does not enable explicit selection of viewpoint-relevant details. To address this limitation, we additionally design a Multi-View Feature Bank that organizes the multi-view prior into a searchable knowledge base for view-adaptive retrieval. The multi-view information is injected through two complementary paths:

\paragraph{Identity-Preserving Latent Injection.}

To provide a direct appearance reference, a shared visual encoder encodes the source video sequence $\mathcal{V}_{\mathrm{src}}$ and the rendered multi-view image set $\mathcal{R}_{\mathrm{mv}}$. The resulting representations are then concatenated and supplied to the denoising DiT blocks. This path anchors the appearance of the inserted object across frames.

\paragraph{Multi-View Feature Bank.}

To fully exploit the multi-view information, we implement a retrieval mechanism through a Multi-View Feature Bank $\mathcal{F}_{\mathrm{mv}}$. This bank is constructed by encoding the rendered multi-view images into a set of key-value pairs $\mathcal{F}_{\mathrm{mv}}=\{(\mathbf{k}_{n},\mathbf{v}_{n})\}_{n=1}^{N}$, where each pair represents object appearance and geometry from a particular perspective.

In our implementation, the attention block processes three information streams: the latent query $\mathbf{q}$, textual context, and multi-view features. During the forward pass, the model first computes standard attention over context tokens and then injects the Multi-View Feature Bank through a separate cross-attention path. For a latent query $\mathbf{x}\in\mathbb{R}^{B\times L\times D}$, we project it into the multi-head attention space as $\mathbf{q}\in\mathbb{R}^{B\times h\times L\times d/h}$. Retrieval from the feature bank is formulated as:
\begin{equation}
\operatorname{Attention}(\mathbf{Q},\mathbf{K},\mathbf{V})=
\operatorname{Softmax}\left(\frac{\mathbf{Q}\mathbf{K}^{T}}{\sqrt{d}}\right)\mathbf{V}
\end{equation}
where $\mathbf{K}$ and $\mathbf{V}$ are the pre-computed keys and values from $\mathcal{F}_{\mathrm{mv}}$.

This formulation allows each frame to retrieve view-relevant appearance cues dynamically from the complete set of multi-view features, improving robustness to viewpoint changes.

\subsubsection{Quality-Aware Weighting}

When the input single-view image has limited quality, the lifting process can produce suboptimal 3D meshes, including blurred occluded regions and stretched textures. We therefore introduce the following strategies to improve the robustness of the injection module.

\paragraph{Synthesized Artifact Augmentation.}
To reduce sensitivity to noisy 3D renders, we employ Synthesized Artifact Augmentation during training by retaining naturally imperfect Hunyuan3D renderings rather than filtering them out during data preparation. These renderings may contain missing geometry, over-smoothed surfaces, or warped textures. Training on this unfiltered distribution exposes the conditioning modules to realistic reconstruction noise, encouraging the model to treat rendered views as fallible guidance.

\paragraph{Semantic Consistency Check.}

To further control the influence of failed 3D reconstructions, we introduce a Semantic Consistency Check module. This module assesses whether the 3D-derived multi-view prior is semantically aligned with the prompt before using it as guidance.

Specifically, we use a pre-trained CLIP model to extract embeddings for the text prompt $\mathbf{e}_{\mathcal{P}}$ and each rendered multi-view image $\mathbf{e}_{I_{\mathrm{mv}}^{n}}$. We compute normalized cosine similarity with temperature parameter $\tau$:
\begin{equation}
\ell_n=\frac{\langle\mathbf{e}_{\mathcal{P}},\mathbf{e}_{I_{\mathrm{mv}}^{n}}\rangle}{\tau},\quad n=1,\ldots,N
\end{equation}

The overall confidence of the multi-view renderings is represented by the mean consistency score $\overline{S}_{\mathrm{cons}}$, which is mapped to a probability through the sigmoid function $\sigma(\cdot)$:
\begin{equation}
\overline{S}_{\mathrm{cons}}=\sigma\left(\frac{1}{N}\sum_{n=1}^{N}\ell_n\right)
\end{equation}

This score modulates the reference scale $\alpha_{\mathrm{ref}}$. Given a base multi-view reference scale $\alpha_0$, the modulated scale is defined as:
\begin{equation}
\alpha_{\mathrm{ref}}=\alpha_0\cdot\overline{S}_{\mathrm{cons}}
\end{equation}

To address localized failures, such as artifacts that occur only at specific viewpoints, we use the per-view consistency score $S_{\mathrm{cons}}^{n}=\sigma(\ell_n)$ to scale individual feature tokens in the Multi-View Feature Bank. For each viewpoint $n$, the corresponding keys $\mathbf{K}_{\mathrm{ref}}^{n}$ and values $\mathbf{V}_{\mathrm{ref}}^{n}$ are scaled as follows:
\begin{equation}
\widetilde{\mathbf{K}}_{\mathrm{ref}}^{n}=S_{\mathrm{cons}}^{n}\mathbf{K}_{\mathrm{ref}}^{n},\quad
\widetilde{\mathbf{V}}_{\mathrm{ref}}^{n}=S_{\mathrm{cons}}^{n}\mathbf{V}_{\mathrm{ref}}^{n}
\end{equation}

This token-level scaling causes the cross-attention mechanism to de-emphasize viewpoints with low semantic alignment. The framework can therefore filter noisy or distorted renders while retaining information from more reliable perspectives.

The final integrated hidden state $\mathbf{x}$ in a transformer block is formulated as:
\begin{equation}
\mathbf{x}=\mathbf{x}_{\mathrm{ctx}}+\alpha_{\mathrm{ref}}\cdot
\operatorname{Attention}\left(\mathbf{Q},\{(\widetilde{\mathbf{K}}_{\mathrm{ref}}^{n},\widetilde{\mathbf{V}}_{\mathrm{ref}}^{n})\}_{n=1}^{N}\right)
\end{equation}
where $\{(\widetilde{\mathbf{K}}_{\mathrm{ref}}^{n},\widetilde{\mathbf{V}}_{\mathrm{ref}}^{n})\}_{n=1}^{N}$ denotes the scaled collection of all multi-view feature tokens. Applying $\alpha_{\mathrm{ref}}$ reduces the influence of the multi-view prior when reconstruction quality is low and encourages greater reliance on textual guidance. The resulting global and token-level weighting improves robustness to reconstruction artifacts and ambiguities in occluded regions.

\subsection{Integration-Aware Consistency Module}

To improve integration between the inserted object and the dynamic background, we introduce the Integration-Aware Consistency Module. This module jointly promotes depth relationships, boundary clarity, and temporal continuity.

\subsubsection{Geometric Grounding}

Inspired by MagicMotion~\cite{MagicMotion}, we append a Depth Head and a Contour Head to the core diffusion architecture. Both heads use a lightweight Feature Pyramid Network (FPN)~\cite{kirillov2019panoptic} to aggregate multi-scale features from the denoising DiT blocks.

Rather than predicting high-resolution images in pixel space, which would require expensive decoding during training, these two heads operate directly in latent space. Predicting latent depth and contour maps enables a more memory-efficient training pipeline.

The Depth Head is supervised by a pre-trained Video Depth Anything model~\cite{video_depth_anything}, which provides pseudo-ground-truth depth maps during training. For each frame $t$, we compute latent depth loss $\mathcal{L}_{\mathrm{depth}}$ as the $\ell_1$ distance between predicted latent depth map $\widehat{\mathbf{Z}}_{d}^{t}$ and pseudo-ground-truth latent depth map $\mathbf{Z}_{d}^{t}$:
\begin{equation}
\mathcal{L}_{\mathrm{depth}}=\frac{1}{T}\sum_{t=1}^{T}\|\widehat{\mathbf{Z}}_{d}^{t}-\mathbf{Z}_{d}^{t}\|_1
\end{equation}

This supervision encourages the model to capture depth ordering and occlusion relationships, thereby improving the geometric grounding of the inserted object within the scene.

The Contour Head sharpens object boundaries. We use a mean squared error (MSE) loss $\mathcal{L}_{\mathrm{seg}}$ to minimize the discrepancy between predicted latent segmentation mask $\widehat{\mathbf{Z}}_{s}^{t}$ and ground-truth mask latent $\mathbf{Z}_{s}^{t}$:
\begin{equation}
\mathcal{L}_{\mathrm{seg}}=\frac{1}{T}\sum_{t=1}^{T}\|\widehat{\mathbf{Z}}_{s}^{t}-\mathbf{Z}_{s}^{t}\|_2^2
\end{equation}

By explicitly supervising the object's spatial extent in latent space, this segmentation objective helps reduce halo artifacts and color bleeding.

\subsubsection{Temporal Consistency Optimization}

To reduce inter-frame flickering and improve motion continuity in the generated video sequence $\mathcal{V}_{\mathrm{gen}}$, we introduce temporal consistency loss $\mathcal{L}_{\mathrm{temp}}$ during fine-tuning. At each fine-tuning iteration, the frozen VAE decodes the predicted clean latent sequence into $\widehat{\mathcal{V}}_0=\{\widehat{I}_0^{t}\}_{t=1}^{T}$. For adjacent decoded frames $(\widehat{I}_0^{t},\widehat{I}_0^{t+1})$, we estimate forward optical flow $\mathbf{F}^{t\rightarrow t+1}$ in the grayscale domain using the Lucas--Kanade approximation.

A differentiable bilinear operator $\mathcal{W}$ warps each decoded prediction toward its subsequent frame. The loss is the mean squared error across adjacent decoded training frames:
\begin{equation}
\mathcal{L}_{\mathrm{temp}}=\frac{1}{(T-1)\cdot3HW}\sum_{t=1}^{T-1}
\left\|\mathcal{W}(\widehat{I}_0^{t},\mathbf{F}^{t\rightarrow t+1})-\widehat{I}_0^{t+1}\right\|_2^2
\end{equation}

The total training objective is a weighted combination of the primary diffusion denoising loss and the auxiliary geometric and temporal constraints:
\begin{equation}
\mathcal{L}_{\mathrm{total}}=\mathcal{L}_{\mathrm{diff}}+\lambda_d\mathcal{L}_{\mathrm{depth}}+\lambda_s\mathcal{L}_{\mathrm{seg}}+\lambda_t\mathcal{L}_{\mathrm{temp}}.
\end{equation}

\section{Experiment}

\subsection{Experimental Setup}

\noindent\textbf{Implementation Details.} Our implementation builds on the pre-trained Wan2.1 VACE 1.3B architecture~\cite{wan2025,vace}, with training conducted on eight NVIDIA H800 GPUs.

Each training clip contains $T=81$ frames and is conditioned on $N=4$ multi-view reference images. We set the base reference scale to $\alpha_0=0.2$ and the loss weights to $\lambda_d=10^{-3}$, $\lambda_s=10^{-3}$, and $\lambda_t=5\times10^{-2}$. LoRA uses a rank of 32 and a learning rate of $10^{-4}$. Additional optimization and inference details are provided in the supplementary material.

\noindent\textbf{Datasets.} Combining YouTube-VIS~\cite{yang2019youtubevis}, DAVIS 2017~\cite{perazzi2016davis}, and MOSE~\cite{MOSE, MOSEv2} yields approximately 41k training clips spanning diverse object categories and motion patterns.

Evaluation is conducted on test videos from DAVIS, VIPSeg~\cite{miao2022vipseg}, and MagicBench~\cite{MagicMotion}.

\noindent\textbf{Evaluation Metrics.}
Our evaluation covers reconstruction fidelity, controllability, and generative quality.
We report PSNR, SSIM, LPIPS, and FVD. Following MagicMotion~\cite{MagicMotion}, spatial controllability is measured using Mask\_IoU and Box\_IoU between the generated outputs and their guidance signals.
Perceptual video quality is assessed across eight dimensions of VBench~\cite{huang2024vbench}. For insertion-specific analysis, Background Consistency and Subject Consistency are computed within the corresponding masked regions.

\subsection{Main Results}
We compare the proposed method with six publicly available video editing or video object insertion methods. We report quantitative and qualitative comparisons.

\noindent\textbf{Quantitative comparison.} To ensure fair comparison across diverse architectures, we adopt a standardized evaluation protocol tailored to the input requirements of each baseline. For methods that rely on first-frame-guided temporal propagation \cite{ku_anyv2v, gao2025loraedit, MagicMotion}, we provide the ground-truth first frame of the target sequence as an edited reference to evaluate their propagation capability. For methods that support single-image reference conditioning \cite{vace, gao2026pisco}, we provide an object rendering from a viewpoint matching that of the ground-truth first frame of the target sequence. This rendering is derived from the same 3D mesh used by the proposed framework, ensuring identical texture and geometry across methods.
In addition to this matched viewpoint, the proposed framework uses three additional rendered views to populate the Multi-View Feature Bank. This setup isolates the contribution of multi-view information relative to baseline methods that use a single view or only the first frame.

As reported in Tab.~\ref{tab:table_main_davis}, the proposed framework achieves strong reconstruction fidelity and perceptual quality. Under bounding-box control, it obtains a PSNR of 23.22 and an SSIM of 0.9026, exceeding the corresponding results of VACE. It also yields the lowest LPIPS and FVD values among the evaluated methods. These results indicate higher similarity to the reference objects and improved temporal realism in generated videos.

For spatial precision, the proposed framework achieves the highest Mask\_IoU and Box\_IoU under bounding-box guidance in Tab.~\ref{tab:table_main_davis}. It also obtains the highest B\_IoU on VIPSeg and MagicBench. These spatial alignment results indicate that generated objects are more accurately localized within the specified control regions throughout the video sequence.

To assess the practical utility of the pipeline, we evaluate Ours (t2i), which implements a complete text-driven object insertion workflow. We first use Qwen-VL-Plus~\cite{Qwen-VL} to generate a detailed textual description of the target object from its appearance in the initial frame. Qwen-Image-2.0-Pro~\cite{wu2025qwenimagetechnicalreport} then synthesizes a reference image from this prompt, and Hunyuan3D 2.0~\cite{hunyuan3d22025tencent} lifts the image into 3D. Despite the domain shift introduced by this multi-stage process, Ours (t2i) remains competitive, achieving a Box\_IoU of 0.7405 and a PSNR of 21.27. It exceeds several baselines that use ground-truth single-view references, indicating that the framework can use multi-view information derived from synthetic 3D meshes effectively.

Results on VIPSeg and MagicBench in Tab.~\ref{tab:table_main} and Tab.~\ref{tab:table_vbench} further support these findings. For reference-free perceptual metrics, the proposed method achieves the highest Subject Consistency and Temporal Flickering scores. These results suggest improved identity preservation and smoother inter-frame transitions across diverse dynamic scenes.

\begin{table}[t]
\centering
\caption{Results on DAVIS 2017. M\_IoU and B\_IoU denote Mask IoU and Box IoU, respectively. ``Ours'' uses a provided reference image; ``Ours (t2i)'' generates this image from text before applying the same insertion pipeline.}
\vspace{-3ex}
\label{tab:table_main_davis}
\resizebox{\linewidth}{!}{
\begin{tabular}{c|c|cccccc}
\toprule
Control Signal & Method & PSNR ($\uparrow$) & SSIM ($\uparrow$) & LPIPS ($\downarrow$) & FVD ($\downarrow$) & M\_IoU ($\uparrow$) &  B\_IoU ($\uparrow$) \\
\midrule
\multirow{2}{*}{-} 
    & AnyV2V~\cite{ku_anyv2v} & 17.87 & 0.5089 & 0.3618 & 1084.92 & - & - \\
    & UniVideo~\cite{wei2025univideo} & 14.86 & 0.4626& 0.3824 & 1822.27 & - & - \\ %
\midrule
\multirow{6}{*}[0pt]{BBox} 
    & VACE (Wan1.3B)~\cite{vace} & 22.85 & \underline{0.9016} & \underline{0.0803} & 957.14 & 0.4090 & 0.6305\\ 
    & MagicMotion (Wan1.3B)~\cite{MagicMotion} & 20.90 & 0.7793 & 0.1571 & 987.97 & 0.4468 & 0.6966\\ 
    & LoRA-Edit~\cite{gao2025loraedit} & 19.60 & 0.6258 & 0.2580 & 827.63 & 0.4058 & 0.6034\\ 
    & PISCO~\cite{gao2026pisco} & 19.13 & 0.8635 & 0.1556 & 843.62 & 0.4542 & 0.6757\\
    & \textbf{Ours} & 23.22 & \textbf{0.9026} & \textbf{0.0769} & 756.19 & 0.5089 & 0.7603 \\ 
    & \textbf{Ours (t2i)}  & 21.27 & 0.7770 & 0.1593 & 894.17 & 0.4751 & 0.7405\\ 
\midrule
\multirow{3}{*}{Mask} 
    & VACE (Wan1.3B) & 23.62 & 0.8138 & 0.1232 & \underline{507.31} & 0.8134 & 0.8644 \\ 
    & MagicMotion (Wan1.3B) & \textbf{23.73} & 0.8150 & 0.1226 & 542.21 & \textbf{0.8254} & \underline{0.8709}\\ 
    & \textbf{Ours} & \underline{23.68} & 0.8132 & 0.1220 & \textbf{460.08} & \underline{0.8208} & \textbf{0.8740} \\ 
\bottomrule
\end{tabular}
}
\vspace{-1ex}
\end{table}

\begin{table*}[t]
\centering
\caption{Reference-based results on VIPSeg and MagicBench. All reference-guided methods receive the same object rendering from the original viewpoint; the proposed method additionally uses three rendered views. B\_IoU denotes Box IoU.}
\vspace{-3ex}
\label{tab:table_main}
\tiny
\resizebox{\linewidth}{!}{
    \begin{tabular}{c|c|ccccc|ccccc} 
    
    \toprule
    \multirow{2}{*}[-2pt]{Control Signal} & \multirow{2}{*}[-2pt]{Method} & \multicolumn{5}{c|}{VIPSeg} & \multicolumn{5}{c}{MagicBench} \\
    
    \cmidrule (lr){3-7} \cmidrule (l){8-12}

    & & PSNR ($\uparrow$) & SSIM ($\uparrow$) & LPIPS ($\downarrow$) & FVD ($\downarrow$) &  B\_IoU ($\uparrow$) 
    
    & PSNR ($\uparrow$) & SSIM ($\uparrow$) & LPIPS ($\downarrow$) & FVD ($\downarrow$) &  B\_IoU ($\uparrow$) \\
    
    \midrule
    \multirow{2}{*}{-} 
    & AnyV2V 
    & {22.93} & {0.8414} & {0.1314} & {816.26}  & - 
    & \underline{22.17} & 0.8291 & \underline{0.0789} & \underline{322.98} & - \\
    & UniVideo
    & 11.96 & 0.4462 & 0.4348 & 2270.05  & -
    & 15.36 & 0.7021 & 0.2521 & 634.24 & - \\
    \midrule
    \multirow{5}{*}{BBox} 
    & VACE (Wan1.3B)
    & 19.84 & 0.8849 & 0.1517 & 831.68 & 0.8053 
    & 21.56 & \underline{0.9295} & 0.0862 & 374.78 & 0.7535 \\
    
    & MagicMotion (Wan1.3B)
    & \underline{23.47} & \underline{0.8892} & \underline{0.1271} & 743.20 & \underline{0.8210} 
    & 21.58 & 0.9293 & 0.0827 & 377.54 & \underline{0.7726}\\
    
    & LoRA-Edit
    & {21.41} & {0.8348} & {0.1444} & {924.76} & {0.7781} 
    & 21.80 & 0.8675 & 0.1514 & 411.41 & 0.7045 \\

    & PISCO~\cite{gao2026pisco}
    & 22.00 & 0.8579 & 0.1255 & \underline{707.42} & 0.5971
    & 22.01 & 0.8285 & 0.1006 & 416.25 & 0.5799 \\
    
    & {Ours}
    & \textbf{24.68} & \textbf{0.8933} & \textbf{0.1056} & \textbf{642.54} & \textbf{0.8250}
    & \textbf{23.07} & \textbf{0.9483} & \textbf{0.058} & \textbf{307.98}  & \textbf{0.7787} \\
    \bottomrule
    \end{tabular}
}
\end{table*}

\begin{table*}[t]
\centering
\caption{Quantitative comparison on VIPSeg and MagicBench using reference-free perceptual metrics.}
\vspace{-3ex}
\label{tab:table_vbench}
\resizebox{\linewidth}{!}{
    \begin{tabular}{c|c|cccccccc|cccccccc} 
    \toprule
    \multirow{2}{*}[-5pt]{Control Signal} & \multirow{2}{*}[-5pt]{Method} & \multicolumn{8}{c|}{VIPSeg} & \multicolumn{8}{c}{MagicBench} \\
    
    \cmidrule (lr){3-10} \cmidrule (l){11-18}
    
    & & \makecell{Subject\\Consistency} ($\uparrow$)
    & \makecell{Background\\Consistency} ($\uparrow$)
    & \makecell{Aesthetic\\Quality} ($\uparrow$)
    & \makecell{Imaging\\Quality} ($\uparrow$)
    & \makecell{Overall\\Consistency} ($\uparrow$)
    & \makecell{Temporal\\Flickering} ($\uparrow$)
    & \makecell{Motion\\Smoothness} ($\uparrow$)
    & \makecell{Dynamic\\Degree} ($\uparrow$) 
    
    & \makecell{Subject\\Consistency} ($\uparrow$)
    & \makecell{Background\\Consistency} ($\uparrow$)
    & \makecell{Aesthetic\\Quality} ($\uparrow$)
    & \makecell{Imaging\\Quality} ($\uparrow$)
    & \makecell{Overall\\Consistency} ($\uparrow$)
    & \makecell{Temporal\\Flickering} ($\uparrow$)
    & \makecell{Motion\\Smoothness} ($\uparrow$)
    & \makecell{Dynamic\\Degree} ($\uparrow$) \\
    \midrule
    
    \multirow{2}{*}{-} & AnyV2V
    & 0.9052 & 0.9266 & \underline{0.5260} & \textbf{72.4} & 0.2189 & 0.9554 & 0.9693 & 0.5135
    & 0.9291 & 0.9223 & \textbf{0.5562} & \textbf{69.4} & \underline{0.1891} & 0.9793 & 0.9803 & 0.3056 \\
    & UniVideo
    & 0.9241 & 0.9206 & 0.5020 & 67.3 & 0.2081 & 0.9675 & 0.9605 & 0.4235 
    & 0.9423 & 0.9317 & 0.5361 & 64.3 & 0.1791 & 0.9743 & 0.9715 & 0.3632 \\
    \midrule
    \multirow{4}{*}{BBox} & VACE (Wan1.3B)
    & 0.9227 & \textbf{0.9298} & 0.5200 & 67.6 & \underline{0.2205} & \underline{0.9675} & \underline{0.9839} & \underline{0.7333} 
    & 0.9402 & 0.9267 & 0.5280 & \underline{65.6} & \textbf{0.1981} & 0.9838 & 0.9874 & \underline{0.6944} \\
    & MagicMotion (Wan1.3B)
    & \underline{0.9269} & 0.9250 & 0.5243 & 65.5 & \textbf{0.2207} & 0.9667 & 0.9838 & 0.7000 
    & \underline{0.9463} & 0.9327 & 0.5219 & 64.6 & 0.1890 & 0.9715 & 0.9751 & 0.5833 \\
    & LoRA-Edit
    & 0.9175 & 0.9213 & 0.4877 & 66.2 & 0.1564 & 0.9628 & 0.9806 & \textbf{0.9195} 
    & 0.9445 & \underline{0.9353} & 0.5180 & 63.7 & 0.1514 & \underline{0.9841} & \underline{0.9880} & \textbf{0.7917} \\
    & \textbf{Ours}
    & \textbf{0.9301} & \underline{0.9279} & \textbf{0.5296} & \underline{67.9} & 0.2183 & \textbf{0.9726} & \textbf{0.9850} & 0.7000
    & \textbf{0.9527} & \textbf{0.9363} & \underline{0.5339} & 64.8 & 0.1889 & \textbf{0.9865} & \textbf{0.9882} & 0.6389 \\
    
    \bottomrule
    \end{tabular}
}
\end{table*}

\noindent\textbf{Qualitative comparison.}
\begin{figure*}[t]
    \centering
    \begin{subfigure}{0.48\linewidth}
        \centering
        \includegraphics[width=\linewidth]{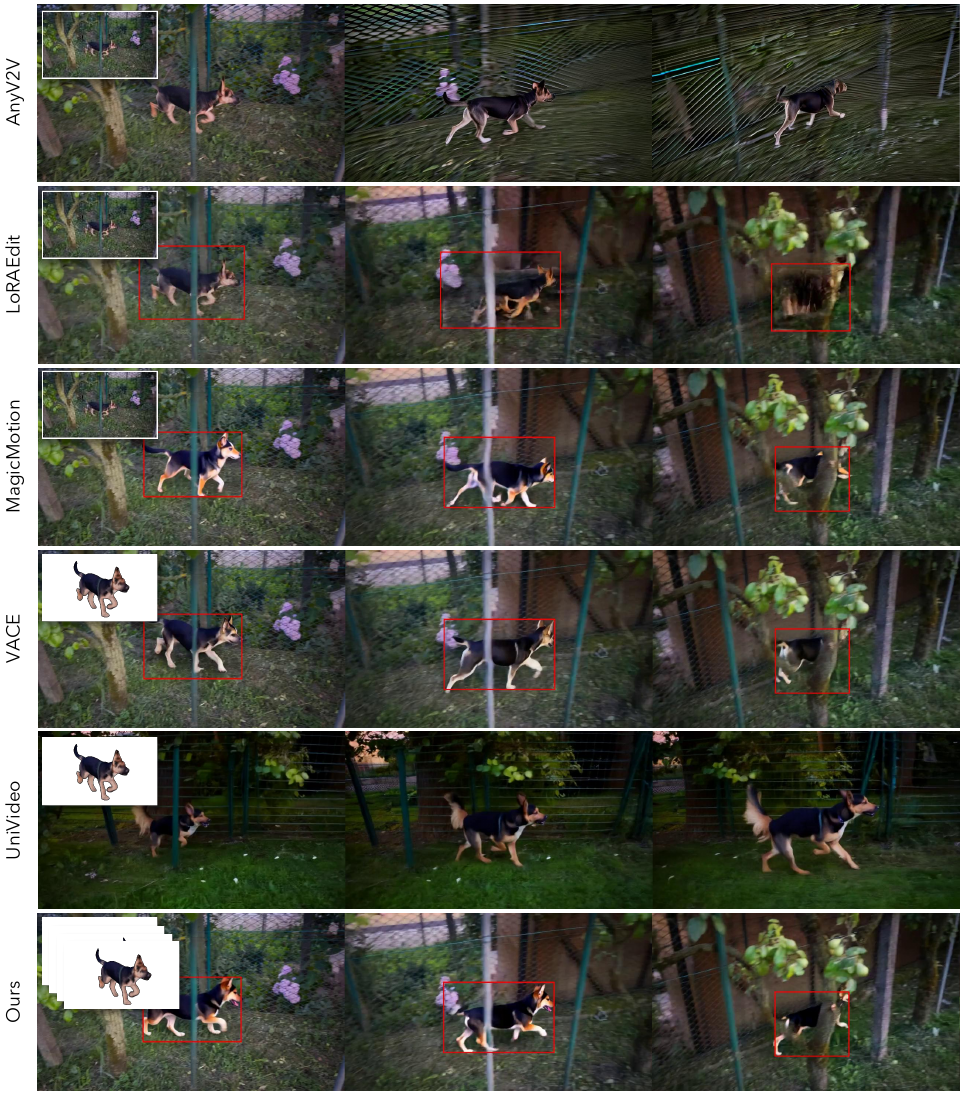}
        \vspace{-4ex}
        \caption{ }
        \label{subfig:qualitative_davis}
    \end{subfigure}\hfill
    \begin{subfigure}{0.48\linewidth}
        \centering
        \includegraphics[width=\linewidth]{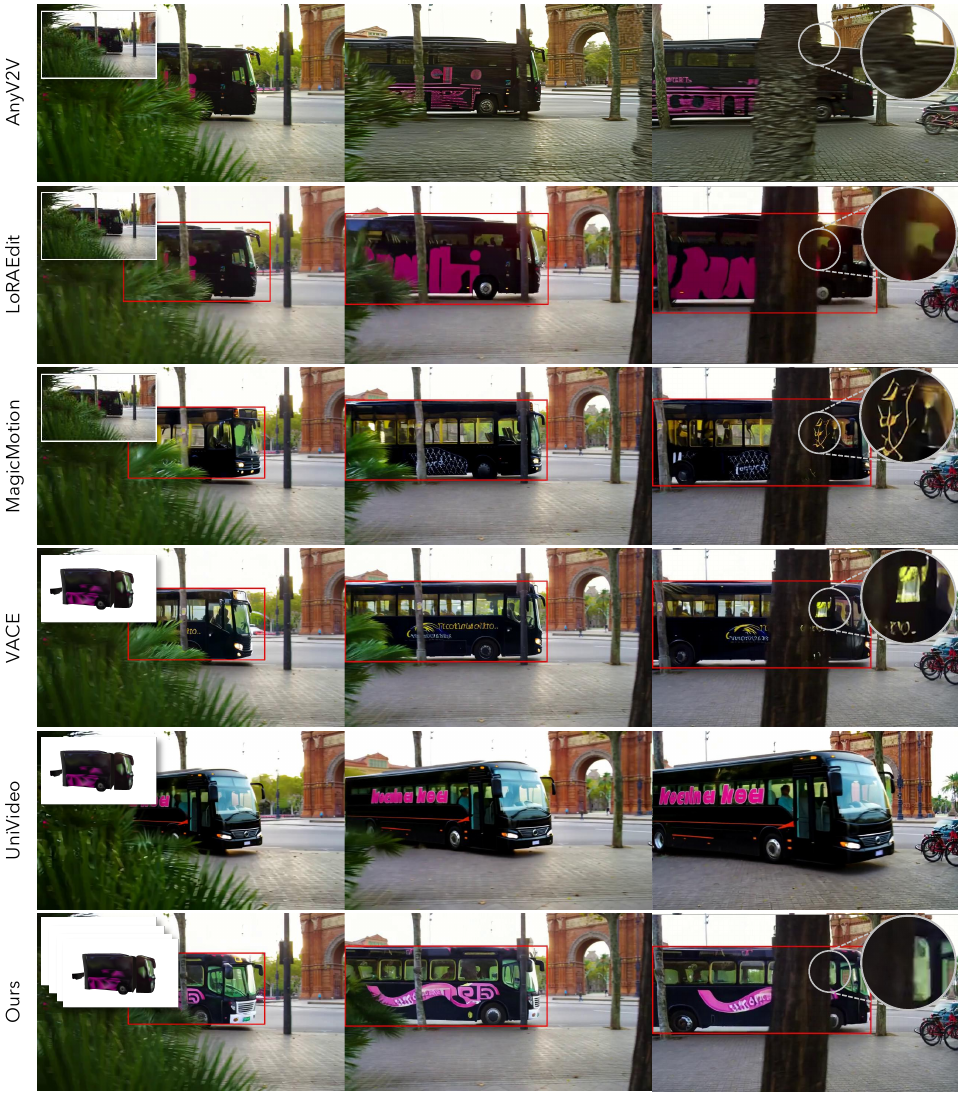}
        \vspace{-4ex}
        \caption{ }
        \label{subfig:qualitative_vipseg}
    \end{subfigure}
    \vspace{-3ex}
    \caption{Qualitative comparison with previous methods. Reference images are shown in the upper left, and red bounding boxes indicate the control signals. Source frames: (a) \textcopyright\ 2017 DAVIS authors, CC BY-NC 4.0~\cite{perazzi2016davis}; (b) \textcopyright\ devinsupertramp, used with permission; source frames accessed through VIPSeg~\cite{miao2022vipseg}.}
    \label{fig:qualitative_main}
    \vspace{-1ex}
\end{figure*}

\begin{table*}[t]
\centering
\caption{Ablation study.}
\vspace{-3ex}
\label{tab:table_ablation}
\tiny
\resizebox{\linewidth}{!}{
\begin{tabular}{c|cccccccc}
\toprule
Method & PSNR ($\uparrow$) & SSIM ($\uparrow$) & LPIPS ($\downarrow$) & FVD ($\downarrow$) & M\_IoU ($\uparrow$) &  B\_IoU ($\uparrow$) & \makecell{Subject\\Consistency} ($\uparrow$) & \makecell{Temporal\\Flickering} ($\uparrow$) \\
\midrule

w/o multi-view prior & 21.45 & 0.7809 & 0.1526 & 826.80 & {0.5003} & \underline{0.7424} & 0.8836 & 0.9307 \\
w/o identity-preserving latent injection & {22.73} & {0.9014} & {0.0797} & 1008.46 & 0.4259 & 0.6593 & 0.8832 & 0.9291 \\
w/o multi-view feature bank & 21.48 & 0.7816 & 0.1511 & \underline{779.02} & \underline{0.5063} & 0.7355 & 0.8833 & 0.9308 \\
w/o semantic consistency check & {23.16} & {0.9013} & {0.0795} & 819.83 & 0.4613 & 0.7361 & 0.8758 & \textbf{0.9322}\\
w/o geometric grounding & \underline{23.20} & \textbf{0.9041} & \underline{0.0795} & 851.47 & 0.4287 & 0.6608 & 0.8817 & \underline{0.9306} \\
w/o temporal consistency optimization & 20.42 & 0.7705 & 0.1604 & 982.17 & 0.4625 & 0.7019 & 0.8746 & 0.9260 \\

\midrule
\textbf{Ours} & \textbf{23.22} & \underline{0.9026} & \textbf{0.0769} & \textbf{756.19} & \textbf{0.5089} & \textbf{0.7603}  &\textbf{ 0.8851} & \textbf{0.9322}\\ 
\bottomrule
\end{tabular}
}
\end{table*}

\begin{figure*}[t]
    \centering
    \begin{subfigure}[t]{0.33\textwidth}
        \centering
        \includegraphics[width=\linewidth]{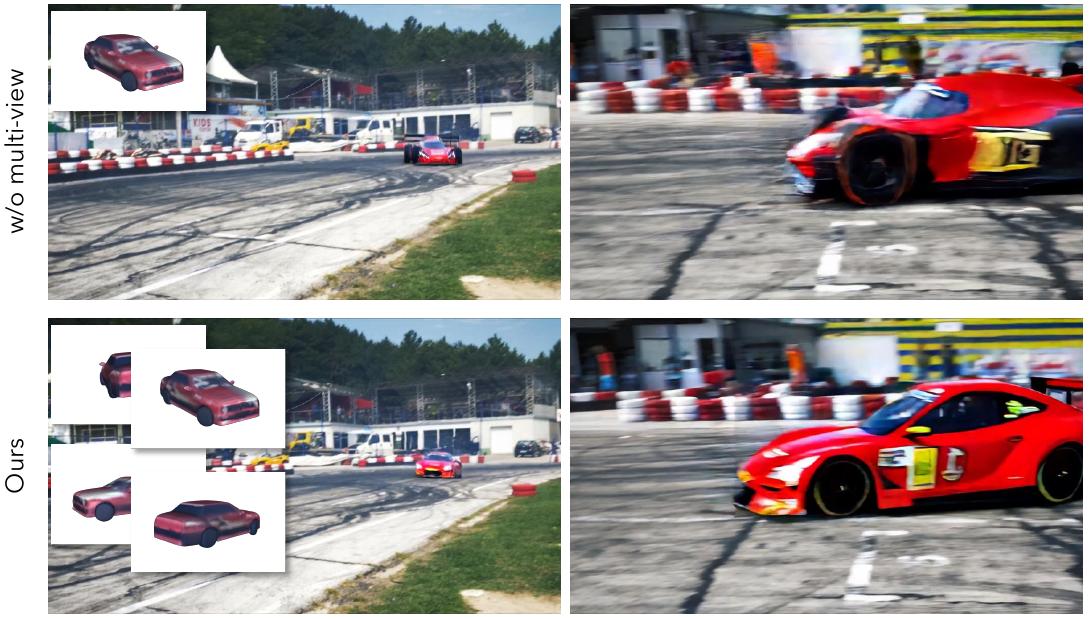}
        \vspace{-4ex}
        \caption{Multi-view prior.}
        \label{fig:ablation multiview}
    \end{subfigure}\hfill
    \begin{subfigure}[t]{0.24\textwidth}
        \centering
        \includegraphics[width=0.9\linewidth]{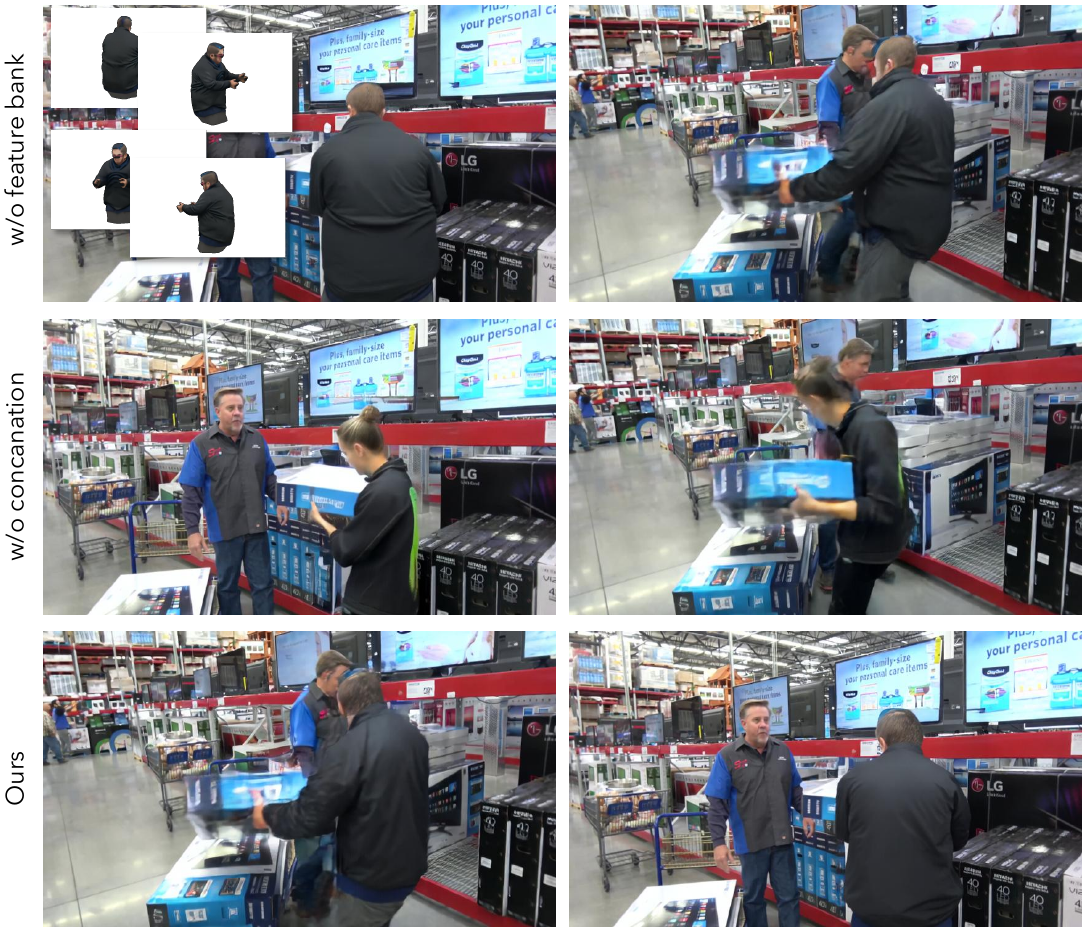}
        \vspace{-1ex}
        \caption{Dual-path reference injection.}
        \label{fig:ablation dual path}
    \end{subfigure}\hfill
    \begin{subfigure}[t]{0.38\textwidth}
        \centering
        \includegraphics[width=\linewidth]{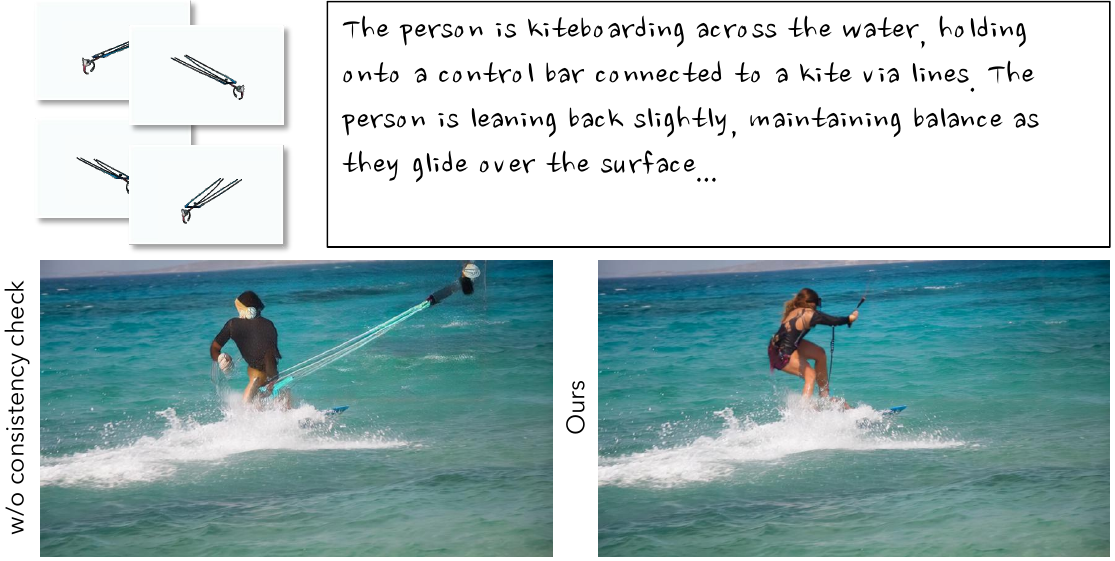}
        \vspace{-4ex}
        \caption{Semantic Consistency Check.}
        \label{fig:ablation conscheck}
    \end{subfigure}
    \vspace{-3ex}
    \caption{Visual ablations. (a) Multi-view priors improve view-consistent appearance. (b) The dual-path design combines structural anchoring with pose-adaptive retrieval. (c) SCC suppresses semantically incomplete rendered priors. Source frames: \textcopyright\ 2017 DAVIS authors, CC BY-NC 4.0~\cite{perazzi2016davis}.}
    \label{fig:visual_ablations}
    \vspace{-1ex}
\end{figure*}

A key requirement of video object insertion is to preserve the original background. Methods without explicit spatial control, such as AnyV2V and UniVideo, can modify regions outside the target area. Under substantial viewpoint changes, this can lead to background blurring, geometric distortion, and structural deformation, as shown in Fig.~\ref{subfig:qualitative_davis}. In contrast, bounding-box control in the proposed framework confines generation to the specified region and better preserves the background.

Realistic integration of new assets requires accurate modeling of depth and occlusion relationships. We examine a challenging scenario in Fig.~\ref{subfig:qualitative_vipseg}, where a bus is inserted into a street scene with foreground trees. AnyV2V and LoRA-Edit produce blurred bus boundaries, MagicMotion and VACE blend bus and tree textures, and UniVideo removes foreground trees. The proposed method retains the trees, places the bus behind them, and produces a sharper boundary.

\subsection{Ablation Studies}

We conduct ablation studies to evaluate the contribution of each core component to multi-view consistency, geometric plausibility, and temporal stability.

\noindent\textbf{Multi-view Prior.}
A central component of the framework is the Multi-view Feature Bank derived from multi-view reference images. We compare the full model with a variant conditioned on only a single reference image.
As illustrated in Fig.~\ref{fig:ablation multiview}, the single-view variant performs reasonably in static scenes but exhibits identity drift and texture stretching under substantial object rotation. Quantitatively, the Subject Consistency score decreases from 0.8851 to 0.8836. This result indicates that the searchable knowledge base provided by $\mathcal{F}_{\mathrm{mv}}$ supports synthesis of view-consistent textures for regions occluded in $I_{\mathrm{ref}}$.

\noindent\textbf{Dual-Path Reference Injection.}
We assess the Dual-Path Reference Injection by separately evaluating the Identity-Preserving Latent Injection path and the Feature Bank path.
To disentangle the benefit of additional multi-view inputs from that of the proposed retrieval architecture, the \textit{w/o multi-view feature bank} variant receives the same four rendered views as the full model but removes the Multi-View Feature Bank retrieval path. The four views are still injected through Identity-Preserving Latent Injection, so the comparison isolates the contribution of view-adaptive feature retrieval under identical multi-view inputs.

The IPLI-only variant conditions the model on concatenated multi-view reference latents without feature-bank retrieval. While this direct conditioning provides strong identity guidance, it cannot adaptively select viewpoint-relevant features during large object transformations. As shown in Fig.~\ref{fig:ablation dual path}, when the foreground undergoes substantial rotation, the static reference information is difficult to align with the evolving target pose, resulting in geometric distortion of the human body.
The MVFB-only variant relies on decoupled cross-attention retrieval without direct latent injection. Its view-adaptive retrieval better accommodates foreground rotation and reduces geometric warping, but provides weaker structural anchoring for identity preservation. As shown in Fig.~\ref{fig:ablation dual path}, the generated person has noticeably slimmer body proportions than the reference subject. Without direct latent guidance from IPLI, feature retrieval alone does not sufficiently constrain the object's shape and proportions.
The full dual-path design combines direct latent anchoring from IPLI with view-adaptive feature retrieval from MVFB. IPLI constrains object shape and identity, whereas MVFB supplies viewpoint-relevant appearance cues during large transformations. Their complementary effects reduce geometric distortion under rotation, as illustrated in Fig.~\ref{fig:ablation dual path}, while the full model's Subject Consistency score further supports its improved identity preservation.

\noindent\textbf{Semantic Consistency Check.}
We assess the effectiveness of SCC in handling imperfect 3D reconstructions. In one failure case, the 3D lifting system produces incomplete renderings for a kiteboarding scene: the multi-view renders contain only the control bar and omit the person. This outcome conflicts with the input prompt, which describes "a person leaning back slightly while kiteboarding."
As shown in Fig.~\ref{fig:ablation conscheck}, the variant without Semantic Consistency Check follows the flawed 3D renders and produces a blurred, indistinct human shape. In contrast, the full model uses CLIP-based alignment scores to detect the semantic mismatch. By reducing the reference scale $\alpha_{\mathrm{ref}}$ and de-emphasizing conflicting multi-view tokens, the model increases its reliance on the text prompt and generates a clear kiteboarding person. Quantitatively, removing SCC increases FVD from 756.19 to 819.83 and decreases Subject Consistency from 0.8851 to 0.8758, supporting the use of SCC for semantically mismatched Hunyuan3D renderings.
\noindent\textbf{Geometry Grounding.}
We evaluate the contribution of the FPN-based Latent Depth and Contour Heads. Removing the Depth Head produces implausible spatial relationships, such as objects overlapping incorrectly with background elements. Omitting the Contour Head produces blurred boundaries and color bleeding, in which foreground pixels contaminate the background. The quantitative results in Tab.~\ref{tab:table_ablation} further support the contribution of Geometry Grounding to spatial alignment.

\noindent\textbf{Temporal Consistency Optimization.}
We evaluate the effect of flow-guided temporal loss $\mathcal{L}_{\mathrm{temp}}$. As indicated by the Temporal Flickering metric in Tab.~\ref{tab:table_ablation}, the Lucas--Kanade-based motion constraint improves the metric from 0.9260 to 0.9322. By explicitly warping decoded prediction $\widehat{I}_0^t$ toward $\widehat{I}_0^{t+1}$ during fine-tuning, the model produces smoother transitions.

\section{Conclusion}

We presented a multi-view-aided framework for controllable video object insertion. In the View-Consistent Conditioning Module, dual-path reference injection combines stable identity anchoring with view-adaptive feature retrieval to preserve the inserted object's identity across viewpoints, while Quality-Aware Weighting reduces sensitivity to imperfect 3D renderings. The Integration-Aware Consistency Module introduces latent depth, latent contour, and temporal objectives to improve layering, boundaries, and inter-frame stability. Experiments on three benchmarks, ablation studies, targeted robustness analyses, and a user study validate the effectiveness of each component and show improvements in visual quality, controllability, identity consistency, and perceptual integration with the background relative to baseline methods.

\begin{acks}
This work was supported by the National Natural Science Foundation of China (Project Number 62595774), Key Laboratory of Intelligent Perception and Human-Machine Collaboration (Shanghaitech University), Ministry of Education, HPC Platform of ShanghaiTech University.
\end{acks}

\newpage

\bibliographystyle{ACM-Reference-Format}
\balance
\bibliography{main}

\begin{thebibliography}{8}

\bibitem{supp:hunyuan3d22025tencent}
Tencent Hunyuan3D Team. 2025. Hunyuan3D 2.0: Scaling Diffusion Models for High Resolution Textured 3D Assets Generation. arXiv:2501.12202.

\bibitem{supp:MagicMotion}
Quanhao Li, Zhen Xing, Rui Wang, Hui Zhang, Qi Dai, and Zuxuan Wu. 2025. MagicMotion: Controllable Video Generation with Dense-to-Sparse Trajectory Guidance. In \textit{Proceedings of the IEEE/CVF International Conference on Computer Vision (ICCV)}. 12112--12123.

\bibitem{supp:miao2022vipseg}
Jiaxu Miao, Xiaohan Wang, Yu Wu, Wei Li, Xu Zhang, Yunchao Wei, and Yi Yang. 2022. Large-scale Video Panoptic Segmentation in the Wild: A Benchmark. In \textit{Proceedings of the IEEE Conference on Computer Vision and Pattern Recognition}.

\bibitem{supp:perazzi2016davis}
Federico Perazzi, Jordi Pont-Tuset, Brian McWilliams, Luc Van Gool, Markus Gross, and Alexander Sorkine-Hornung. 2016. A Benchmark Dataset and Evaluation Methodology for Video Object Segmentation. In \textit{Proceedings of the IEEE Conference on Computer Vision and Pattern Recognition}. 724--732.

\bibitem{supp:teed2020raft}
Zachary Teed and Jia Deng. 2020. RAFT: Recurrent All-Pairs Field Transforms for Optical Flow. In \textit{European Conference on Computer Vision}. Springer, 402--419.

\bibitem{supp:vace}
Zeyinzi Jiang, Zhen Han, Chaojie Mao, Jingfeng Zhang, Yulin Pan, and Yu Liu. 2025. VACE: All-in-One Video Creation and Editing. In \textit{Proceedings of the IEEE/CVF International Conference on Computer Vision}. 17191--17202.

\bibitem{supp:video_depth_anything}
Sili Chen, Hengkai Guo, Shengnan Zhu, Feihu Zhang, Zilong Huang, Jiashi Feng, and Bingyi Kang. 2025. Video Depth Anything: Consistent Depth Estimation for Super-Long Videos. arXiv:2501.12375.

\bibitem{supp:waymo_open_dataset}
Waymo LLC. 2019--2025. Waymo Open Dataset: An Autonomous Driving Dataset. \url{https://www.waymo.com/open}. Waymo Dataset License Agreement for Non-Commercial Use.

\end{thebibliography}

\newpage

\title{Controllable Video Object Insertion via Multi-view Priors: Supplementary Material}

\makeatletter
\let\@mkauthors\relax
\let\@mkteasers\relax
\global\let\@abstract\@lempty
\global\let\@concepts\@empty
\global\let\@keywords\@empty
\gdef\@authornotes{}
\renewcommand{\footnotetextcopyrightpermission}[1]{}
\renewcommand{\footnotetextauthorsaddresses}[1]{}
\makeatother
\settopmatter{printacmref=false}
\renewcommand{\shortauthors}{Xia et al.}
\maketitle


\appendix

\section{Additional Implementation Details}

\subsection{Dataset Pre-processing}

\noindent\textbf{Reference Selection Strategy.} 
To prevent the model from exploiting an overly easy correspondence between a target clip and a reference derived from its first frame, we adopt a \textbf{Temporal Distance Maximization} strategy for constructing reference-target pairs. Specifically, for each $T=81$ target clip sampled from a longer source video, we select $I_{\mathrm{ref}}$ from the frame that is temporally farthest from it. For example, given a 100-frame video, a target clip spanning frames 1--81 uses frame 100 as the reference, whereas a clip spanning frames 20--100 uses frame 1. By preventing the reference rendering from consistently matching the initial viewpoint and appearance of the target clip, this strategy increases the reference-target gap and better reflects practical settings in which a user-provided reference may differ substantially from the target video.

\noindent\textbf{3D Lifting and Rendering.} 
For each selected $I_{\mathrm{ref}}$, Hunyuan3D 2.0~\cite{supp:hunyuan3d22025tencent} reconstructs a textured mesh, from which we render $\mathcal{R}_{\mathrm{mv}}=\{I_{\mathrm{mv}}^{n}\}_{n=1}^{N}$ with $N=4$ uniformly distributed camera views.


\noindent\textbf{Multi-Granularity Captioning.} 
To accommodate user inputs with varying levels of detail, we use Qwen2.5-VL-72B-Instruct to generate four captions for each training sample: two concise summaries and two detailed descriptions. During training, one caption is uniformly sampled from this set to reduce sensitivity to prompt phrasing.

\noindent\textbf{Pseudo-Ground Truth Preparation.} 
To supervise the Integration-Aware Consistency Module, we run Video Depth Anything~\cite{supp:video_depth_anything} on each training clip and encode the resulting pseudo-depth maps to the latent space using the same frozen Wan VAE encoder as the RGB video frames.

\subsection{Fine-tuning Strategy and Inference Configuration}

To preserve the generative priors of the foundation model, we keep the Wan DiT backbone, VAE, and T5 encoder frozen. The image encoder used to process multi-view renderings and the CLIP model used for the Semantic Consistency Check are also frozen.

We attach rank-32 LoRA adapters to the VACE branch of the DiT modules. The Multi-View Feature Bank adapter, Contour Head, and Depth Head are trained from scratch. We use AdamW with a learning rate of $1 \times 10^{-4}$ and a batch size of one clip per GPU. Training takes approximately 72 hours on 8 NVIDIA H800 GPUs.

During inference, we employ 50 sampling steps and set the VACE scale to 1.0. The quality-aware weighting mechanism is enabled to adaptively modulate the influence of rendered priors.

\section{Extended Experiments}

\subsection{Ablation on View Count $N$}

We study how the number of multi-view reference images ($N$) affects reconstruction fidelity and identity consistency. For a controlled comparison, all configurations with $N \geq 2$ use uniform panoramic sampling: the object is rendered from cameras placed at evenly spaced azimuth angles over $360^{\circ}$. Specifically, $N=3$ uses azimuth angles of $0^{\circ}$, $120^{\circ}$, and $240^{\circ}$, whereas our default $N=4$ setting uses $0^{\circ}$, $90^{\circ}$, $180^{\circ}$, and $270^{\circ}$. This sampling scheme provides the Multi-View Feature Bank with complementary observations and maximizes azimuthal coverage of the object for a fixed number of reference views.

As summarized in Table~\ref{tab:table_ablation_view_cound}, the results indicate that increasing $N$ from 1 to 4 leads to a significant improvement in reconstruction fidelity and perceptual quality. Specifically, $N=1$ suffers from the lowest SSIM (0.7809) and a high FVD (826.80). While $N=2$ and $N=3$ progressively improve PSNR and SSIM, our proposed $N=4$ configuration achieves the best \textit{Subject Consistency} (0.8851) and the lowest LPIPS (0.0769), striking an optimal balance between multi-view coverage and computational overhead. 

Interestingly, further increasing $N$ to 8 yields only marginal improvements in PSNR (+0.15 dB) and FVD, yet it introduces a disproportionate increase in computational cost, with FLOPs increasing from 16,219,511M to 26,853,860M (a 65\% increase). Moreover, $N=8$ actually exhibits a slight degradation in \textit{Subject Consistency} compared to $N=4$, suggesting that excessive multiview tokens may introduce redundant or conflicting information that complicates the cross-attention retrieval process. Therefore, $N=4$ is selected as our default setting to ensure a favorable performance-efficiency trade-off.

\begin{table}[thbp]
\caption{Ablation study of the number of views.}
\vspace{-3ex}
\label{tab:table_ablation_view_cound}
\tiny
\resizebox{\linewidth}{!}{
\begin{tabular}{c|cccccc}
\toprule
View Count $N$ & PSNR ($\uparrow$) & SSIM ($\uparrow$) & LPIPS ($\downarrow$) & FVD ($\downarrow$) & \makecell{Subject\\Consistency} ($\uparrow$) & FLOPs (M) \\
\midrule

1 & 21.45 & 0.7809 & 0.1526 & 826.80 & \underline{0.8836} & 9,795,597\\ 
2 &  22.94 & 0.8998 & 0.0818 & {822.82} & 0.8787 & 11,789,107\\
3 &  \underline{23.29} & {0.9021} & {0.0789} & 863.69 & {0.8795} & 13,930,411\\
\textbf{4 (Ours)}& {23.22} & \underline{0.9026} & \textbf{0.0769} & \underline{756.19} & \textbf{0.8851} & 16,219,511\\
8 & \textbf{23.37} & \textbf{0.9028} & \underline{0.0782} & \textbf{743.99} & {0.8823} & 26,853,860 \\

\bottomrule
\end{tabular}
}
\end{table}

\subsection{Ablation on Integration-Aware Consistency Components}

\begin{table*}[t]
\caption{Ablation study of components in the Integration-Aware Consistency Module.}
\vspace{-3ex}
\label{tab:table_more_ablation}
\tiny
\resizebox{\linewidth}{!}{
\begin{tabular}{ccc|cccccccc}
\toprule
\multicolumn{3}{c|}{Component} & \multirow{2}{*}{PSNR ($\uparrow$)} & \multirow{2}{*}{SSIM ($\uparrow$)} & \multirow{2}{*}{LPIPS ($\downarrow$)} & \multirow{2}{*}{FVD ($\downarrow$)} & \multirow{2}{*}{M\_IoU ($\uparrow$)} & \multirow{2}{*}{B\_IoU ($\uparrow$)} & \multirow{2}{*}{\makecell{Subject\\Consistency} ($\uparrow$)} & \multirow{2}{*}{\makecell{Temporal\\Flickering} ($\uparrow$)} \\

CH & DH & TCO & & & & & & & & \\
    \midrule
    &\ding{51} &\ding{51}
    & \underline{23.10} & 0.8911 & 0.0845 & 868.30 & 0.4353 & 0.7101 & \underline{0.8812} & 0.9304 \\
    \ding{51}& &\ding{51}
    & 22.30 & 0.8918 & 0.0793 & 868.30 & 0.4813 & 0.7361 & \textbf{0.8851} & 0.9304 \\
    \ding{51}&\ding{51} &
    & 20.42 & 0.7705 & 0.1604 & 982.17 & 0.4625 & 0.7019 & 0.8746 & 0.9260 \\
    \midrule
    \ding{51}&\ding{51} &\ding{51}
    & \textbf{23.22} & \textbf{0.9026} & \textbf{0.0769} & \textbf{756.19} & \textbf{0.5089} & \textbf{0.7603} & \textbf{0.8851} & \textbf{0.9322}\\
\bottomrule
\end{tabular}
}
\end{table*}

We evaluate the three components of the Integration-Aware Consistency Module: the Contour Head (CH), Depth Head (DH), and Temporal Consistency Optimization (TCO). Table~\ref{tab:table_more_ablation} reports variants that remove one component at a time while keeping the rest of the framework unchanged.

Removing CH decreases Box\_IoU from 0.7603 to 0.7101, confirming its role in sharpening object boundaries and reducing boundary artifacts. Removing DH lowers Box\_IoU to 0.7361 because the model lacks explicit depth-aware reasoning for complex foreground-background layering. Without TCO, FVD increases from 756.19 to 982.17 and Temporal Flickering decreases from 0.9322 to 0.9260, demonstrating its contribution to temporal stability.

\subsection{Robustness to Depth Pseudo-Labels}

Video Depth Anything produces reliable pseudo-depth maps for most training clips. To evaluate robustness to occasional estimation errors, we perturb these pseudo-labels and replace them with predictions from a weaker depth estimator. Both variants remain stable, with moderate degradation rather than collapse, suggesting that the Depth Head benefits from approximate geometric supervision.
Stronger depth models can replace the current choices without changing the architecture.

\begin{table}[t]
\centering
\caption{Sensitivity to pseudo-depth supervision on DAVIS 2017.}
\label{tab:depth_robustness}
\tiny
\resizebox{\linewidth}{!}{
\begin{tabular}{lcccc}
\toprule
Setting & PSNR $\uparrow$ & SSIM $\uparrow$ & LPIPS $\downarrow$ & FVD $\downarrow$ \\
\midrule
Perturbed depth pseudo-labels & 22.75 & 0.8918 & 0.0814 & 810.23 \\
Weaker depth estimator & 23.18 & 0.9016 & 0.0785 & 789.63 \\
Full model & \textbf{23.22} & \textbf{0.9026} & \textbf{0.0769} & \textbf{756.19} \\
\bottomrule
\end{tabular}}
\end{table}

\subsection{Temporal Estimator Analysis}

LK flow provides an efficient temporal estimator when motion between adjacent frames is small. This condition is well matched to our training data: 99.2\% of pixels have displacements of at most two pixels, and the 95th-percentile displacement is 0.75 pixels. As shown in Table~\ref{tab:flow_estimator}, applying the temporal loss with LK substantially improves both FVD and Temporal Flickering compared with removing temporal supervision. RAFT~\cite{supp:teed2020raft} provides further gains but requires more expensive flow estimation during training. We therefore use LK as the default estimator and regard RAFT as a stronger alternative when additional computation is available.

\begin{table}[t]
\centering
\caption{Comparison of temporal estimators on DAVIS 2017.}
\label{tab:flow_estimator}
\small
\begin{tabular}{lcc}
\toprule
Temporal estimator & FVD $\downarrow$ & Temporal Flickering $\uparrow$ \\
\midrule
No temporal loss & 982.17 & 0.9260 \\
LK flow (default) & 756.19 & 0.9322 \\
RAFT flow & \textbf{747.36} & \textbf{0.9372} \\
\bottomrule
\end{tabular}
\end{table}

\subsection{Runtime and Memory}

Table~\ref{tab:runtime_memory} reports inference cost on an NVIDIA A40. Our method is substantially faster than UniVideo and uses 14,102 MB peak memory. The added multi-view retrieval increases cost relative to VACE but remains close to other diffusion-based insertion systems.

\begin{table}[t]
\centering
\caption{Inference efficiency on one NVIDIA A40 GPU. Time is averaged per generated frame.}
\label{tab:runtime_memory}
\small
\begin{tabular}{lcc}
\toprule
Method & Time (s/frame) $\downarrow$ & Peak memory (MB) $\downarrow$ \\
\midrule
UniVideo & 68.0 & 84,238 \\
PISCO & \textbf{6.6} & 31,998 \\
VACE & 7.0 & \textbf{11,948} \\
Ours & 9.4 & 14,102 \\
\bottomrule
\end{tabular}
\end{table}

\subsection{View Relevance Visualization}

\begin{figure}[t]
    \centering
    \includegraphics[width=\linewidth]{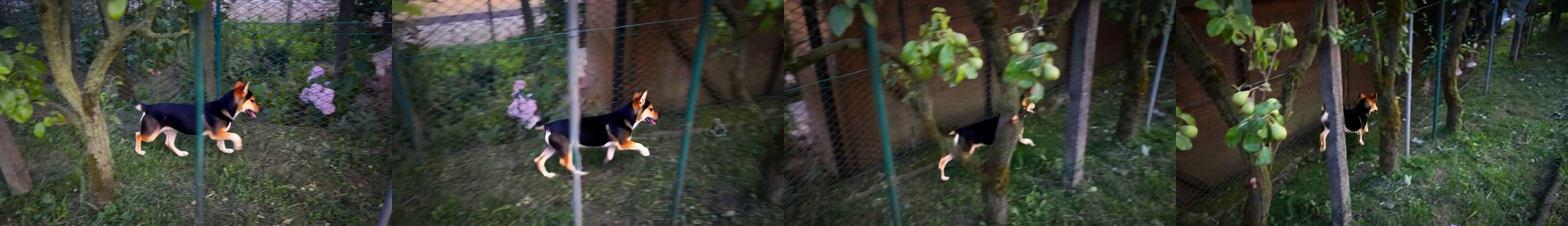}
    \includegraphics[width=\linewidth]{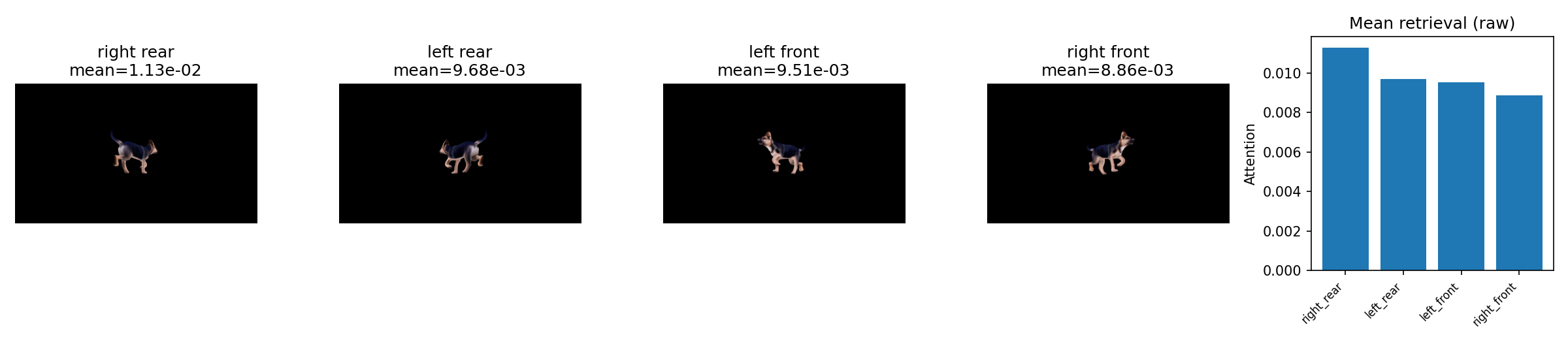}
    \caption{View-relevance analysis. Top: generated sequence. Source frames: \textcopyright\ 2017 DAVIS authors, CC BY-NC 4.0~\cite{supp:perazzi2016davis}. Bottom: the four rendered views and their mean retrieval weights.}
    \label{fig:view_attention}
\end{figure}

We aggregate the Multi-View Feature Bank cross-attention weights over query tokens and denoising blocks for the dog-in-forest sequence. Figure~\ref{fig:view_attention} shows that the right-rear reference receives the highest mean retrieval weight throughout the clip. This agrees with the dominant visible side of the generated dog and illustrates that the feature bank learns non-uniform, pose-relevant view selection rather than averaging all views equally.

\subsection{User Study}

\begin{table}[t]
\caption{User study (1--5, higher is better): Identity Fidelity (IF), Temporal Stability (TS), Physical Integration (PI), Background Fidelity (BF), and Overall Ranking (OVR).}
\vspace{-3ex}
\label{tab:table_user_study}
\tiny
\resizebox{\linewidth}{!}{
\begin{tabular}{c|ccccc}
    \toprule
    Method & IF ($\uparrow$) & TS ($\uparrow$) & PI ($\uparrow$) & BF ($\uparrow$) & OVR ($\uparrow$) \\
    \midrule
    AnyV2V & 4.25 & 4.30 & 4.00 & 3.70 & 3.80 \\
    UniVideo & 3.80 & 3.85 & 3.60 & 4.05 & 3.25 \\
    VACE (Wan1.3B) & \underline{4.55} & \textbf{4.75} & \underline{4.45} & \underline{4.85} & \underline{4.40} \\
    MagicMotion (Wan1.3B) & 4.40 & 4.65 & \underline{4.45} & \underline{4.85} & 4.25 \\
    LoRA-Edit & 3.45 & 3.40 & 3.15 & 3.15 & 2.70 \\
    \textbf{Ours} & \textbf{4.70} & \underline{4.70} & \textbf{4.50} & \textbf{4.95} & \textbf{4.70} \\
    \bottomrule
\end{tabular}
}
\end{table}

We conducted a user study with 20 computer-vision researchers using 30 clips randomly sampled from DAVIS and MagicBench. Participants rated our method and five baselines on a 1--5 scale for Identity Fidelity (IF), Temporal Stability (TS), Physical Integration (PI), Background Fidelity (BF), and Overall Ranking (OVR). PI evaluates layering, occlusion correctness, and boundary naturalness. Because existing benchmarks lack standardized evaluation protocols and dedicated metrics for post-insertion depth ordering and occlusion-boundary quality, we use PI as targeted perceptual measures.

As shown in Table~\ref{tab:table_user_study}, our method ranks first in four of the five criteria. Its IF score of 4.70 indicates strong identity preservation. Although VACE~\cite{supp:vace} and MagicMotion~\cite{supp:MagicMotion} benefit from strong temporal priors, our method remains competitive in TS (4.70) through temporal optimization.

Our method also obtains the highest Overall Ranking (4.70). On the two integration-focused dimensions, it improves the strongest baseline from 4.45 to 4.50 on PI and from 4.85 to 4.95 on BF. These results provide evidence for improved layering, boundary quality, and background preservation beyond generic reconstruction metrics.

\subsection{Diverse Category Insertion}

\begin{figure*}[t]
    \centering
    \begin{subfigure}{0.48\linewidth}
        \centering
        \includegraphics[width=\linewidth]{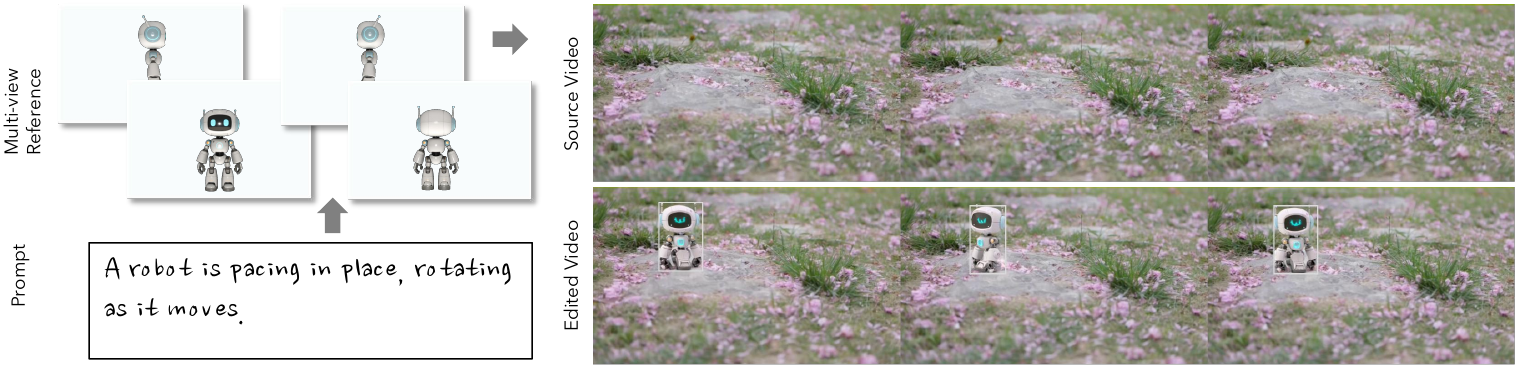}
        \caption{Rigid object insertion.}
        \label{subfig:diverse_rigid}
    \end{subfigure}\hfill
    \begin{subfigure}{0.48\linewidth}
        \centering
        \includegraphics[width=\linewidth]{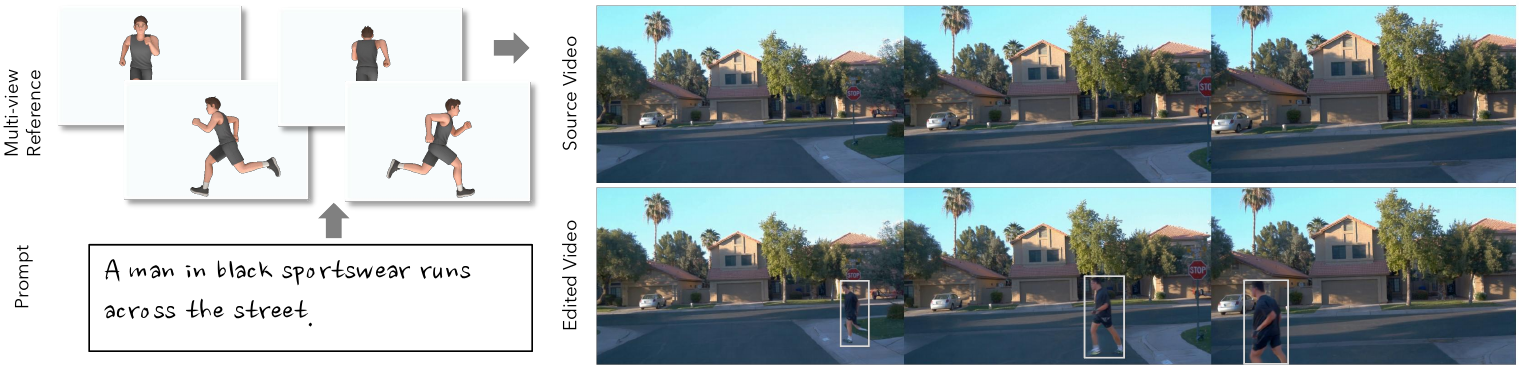}
        \caption{Non-rigid object insertion.}
        \label{subfig:diverse_nonrigid}
    \end{subfigure}
    \\
    \begin{subfigure}{0.48\linewidth}
        \centering
        \includegraphics[width=\linewidth]{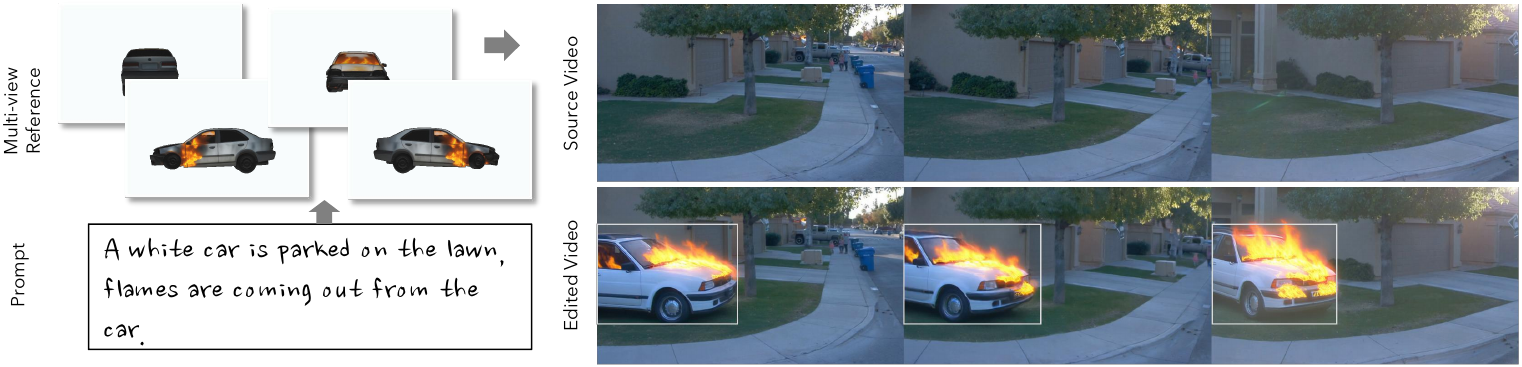}
        \caption{Large camera motion.}
        \label{subfig:diverse_camera}
    \end{subfigure}\hfill
    \begin{subfigure}{0.48\linewidth}
        \centering
        \includegraphics[width=\linewidth]{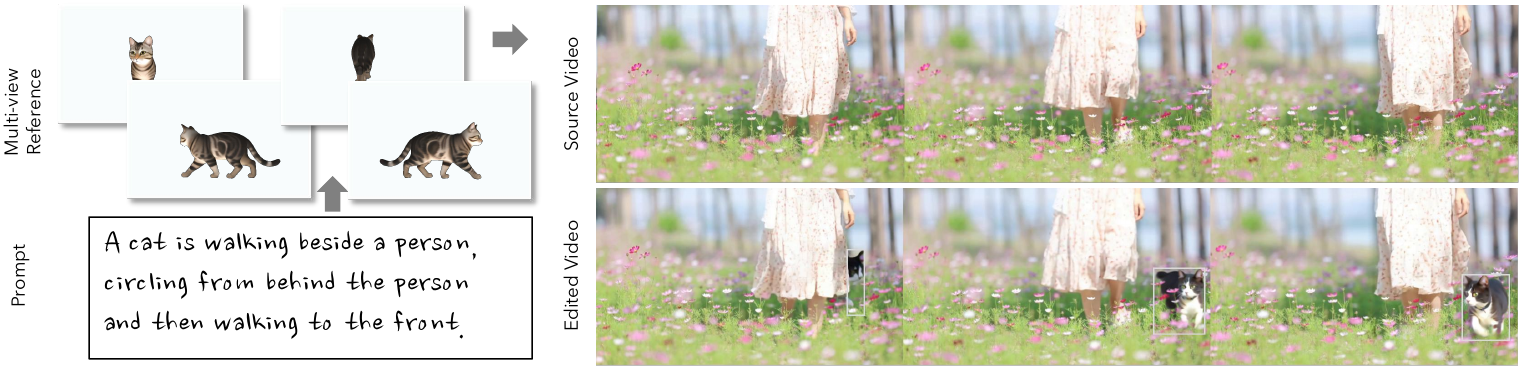}
        \caption{Complex occlusion.}
        \label{subfig:diverse_occlusion}
    \end{subfigure}
    \caption{Results across object types, camera motion, and occlusion conditions. Source frames: (a) and (d) \textcopyright\ Ke Aierxi, used with permission; (b) and (c) \textcopyright\ Waymo LLC, Waymo Open Dataset~\cite{supp:waymo_open_dataset}.}
    \label{fig:diverse_insertion}
\end{figure*}

To further demonstrate the robustness of our framework, we evaluate it across diverse object types and challenging environmental conditions:

\noindent\textbf{Rigid and Non-Rigid Objects.} Figures~\ref{subfig:diverse_rigid} and~\ref{subfig:diverse_nonrigid} illustrate the performance of our framework on rigid and non-rigid targets, respectively. For the rotating robot, our method maintains consistent geometry and appearance across viewpoints while preserving sharp surface details. For the running person, it preserves coherent body proportions and clearly defined limbs throughout the articulated motion. Together, these examples demonstrate robustness to viewpoint changes for rigid objects and articulation for non-rigid subjects.

\noindent\textbf{Large Camera Motion.} As shown in Fig.~\ref{subfig:diverse_camera}, under substantial camera viewpoint changes, our method adapts the appearance of the inserted object to the evolving camera pose, naturally revealing the corresponding object views over time. Throughout this transition, the vehicle maintains coherent geometry and sharp material details without noticeable texture stretching.

\noindent\textbf{Complex Occlusions.} By leveraging the latent depth reasoning, our method achieves realistic layering when inserting objects behind intricate structures like tree branches.

\subsection{Beyond Object Insertion}

\begin{figure*}[t]
    \centering
    \begin{subfigure}{0.48\linewidth}
        \centering
        \includegraphics[width=\linewidth]{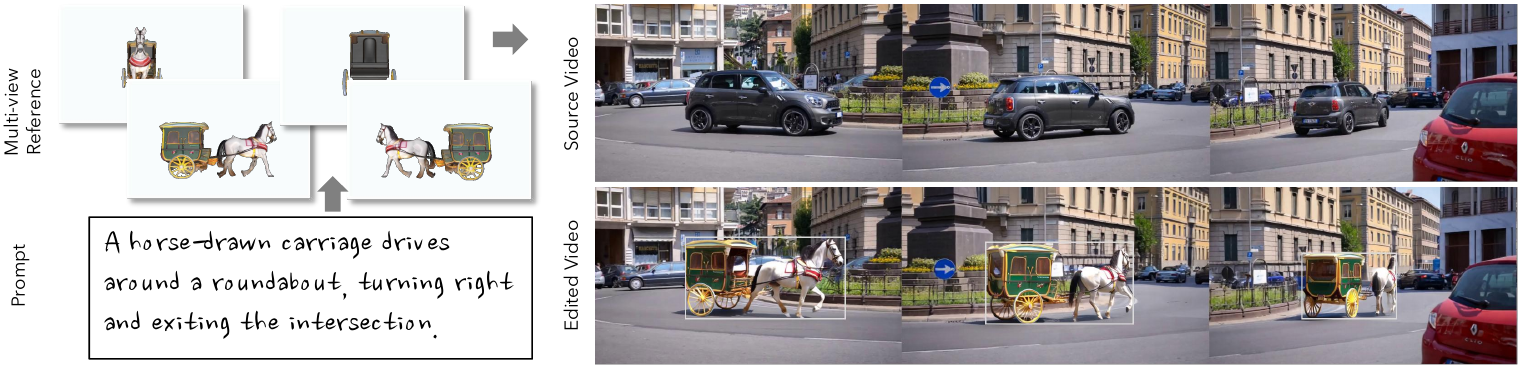}
        \caption{Object Substitution: Replacing a Vehicle with a Carriage.}
        \label{subfig:replace_vehicle}
    \end{subfigure}\hfill
    \begin{subfigure}{0.48\linewidth}
        \centering
        \includegraphics[width=\linewidth]{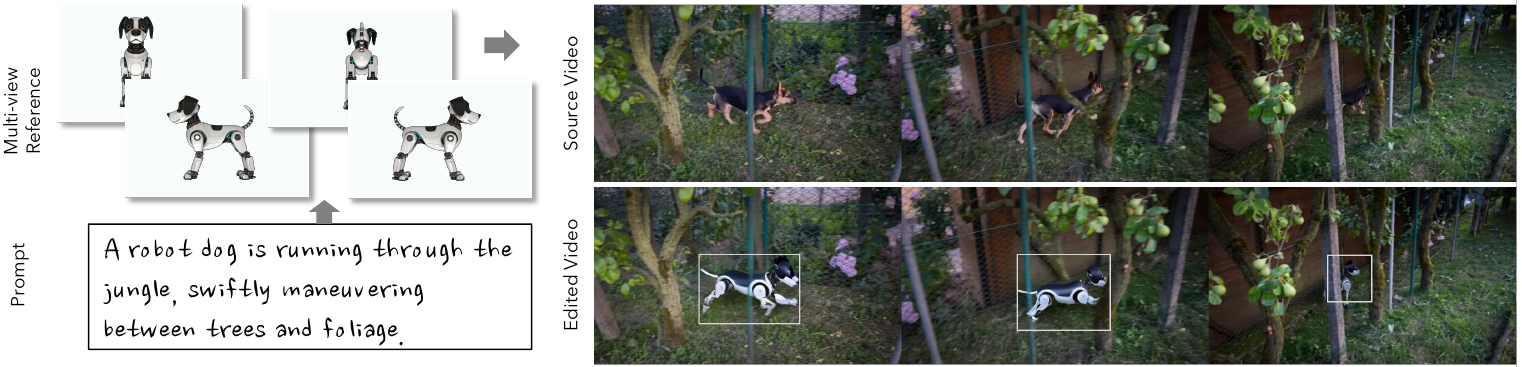}
        \caption{Attribute Manipulation: Transforming a Dog to a Cyber Dog.}
        \label{subfig:replace_attribute}
    \end{subfigure}
    \caption{Broader applications of our framework. Source frames: \textcopyright\ 2017 DAVIS authors, CC BY-NC 4.0~\cite{supp:perazzi2016davis}.}
    \label{fig:broader_application}
\end{figure*}

\begin{figure*}[t]
    \centering
    \begin{subfigure}{0.33\linewidth}
        \centering
        \includegraphics[width=\linewidth]{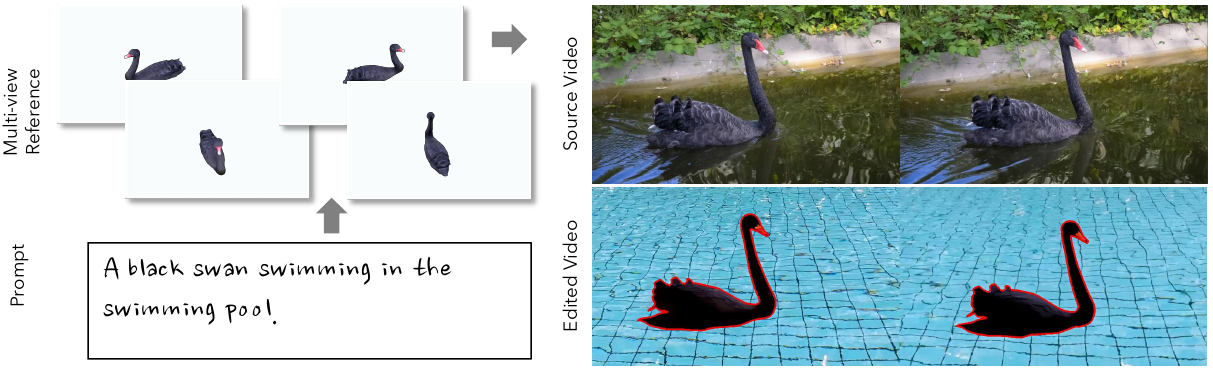}
        \caption{Background Substitution}
        \label{subfig:background_substitution}
    \end{subfigure}\hfill
    \begin{subfigure}{0.33\linewidth}
        \centering
        \includegraphics[width=\linewidth]{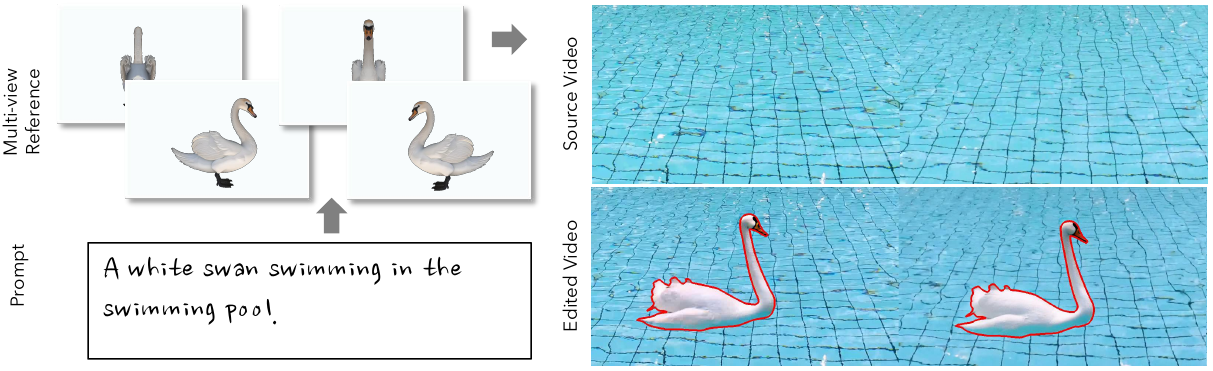}
        \caption{Mask-Controlled Object Insertion}
        \label{subfig:object_insertion_with_mask}
    \end{subfigure}\hfill
    \begin{subfigure}{0.33\linewidth}
        \centering
        \includegraphics[width=\linewidth]{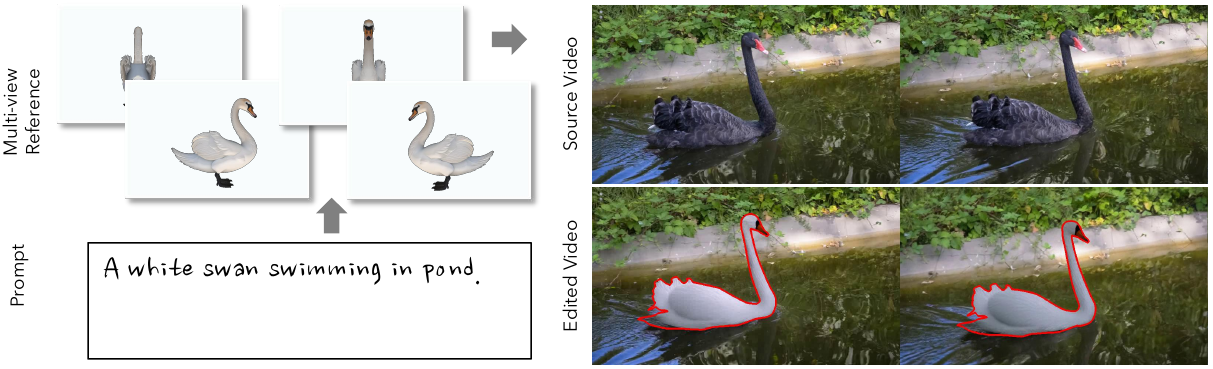}
        \caption{Foreground Substitution}
        \label{subfig:foreground_substitution}
    \end{subfigure}
    \caption{Fine-grained masks enable precise video editing. Source material: \textcopyright\ 2017 DAVIS authors, CC BY-NC 4.0~\cite{supp:perazzi2016davis}}
    \label{fig:mask_control}
\end{figure*}

Beyond insertion, our framework also supports the editing examples in Figs.~\ref{fig:broader_application} and~\ref{fig:mask_control}:

\noindent\textbf{Object Substitution and Attribute Manipulation.} As shown in Fig.~\ref{fig:broader_application}, our approach replaces an existing vehicle with a carriage and transforms a dog into a cyber dog.

\noindent\textbf{Precise Editing.} Our framework supports both coarse bounding-box control and fine-grained mask control. Figure~\ref{fig:mask_control} illustrates background substitution, object insertion, and foreground substitution using fine-grained masks. By providing finer control signals, users can achieve precise modifications without affecting the background or other objects.

\section{Limitations and Future Works}

\subsection{Failure Case Analysis}

\begin{figure}[t]
    \centering
    \includegraphics[width=\linewidth]{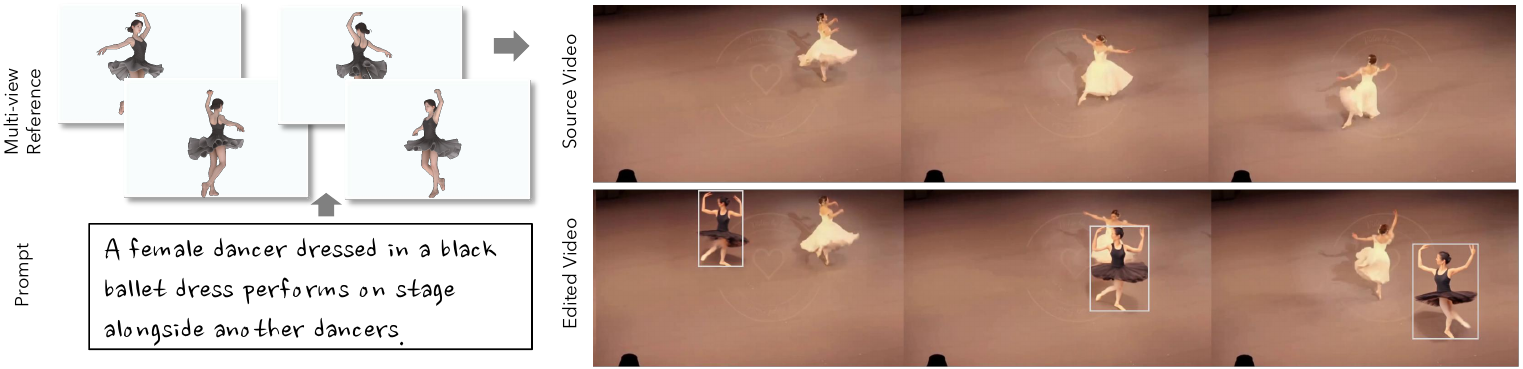}
    \caption{Failure case. A manually specified bounding-box trajectory does not capture the dancer's complex articulation, and the generated motion becomes overly static.}
    \label{fig:failure_case}
\end{figure}

Despite the performance gains, our framework faces several challenges:

\noindent\textbf{Small and Short-Duration Object Insertion.} Our framework is less reliable when the insertion region is spatially small or persists for only a few frames. Because VACE encodes video and mask conditions in a downsampled latent space and injects them as soft hints, small or short-lived masks may produce weak conditioning signals relative to the dominant background context. The model may therefore preserve the original background instead of synthesizing the requested object, leading to weak insertion, blurred textures, or occasional failure.

\noindent\textbf{Complex Physical Interactions.} Generating videos with physically plausible interactions between an inserted character and existing scene objects, such as grasping or complex physical contact, remains challenging. These interactions require reasoning about 3D dynamics, contact constraints, and object responses beyond geometric alignment.

\noindent\textbf{Sensitivity to Control Signal.} If the control signals are spatially or temporally inconsistent with the object's natural behavior, such as providing a fixed-size bounding-box trajectory for a highly deformable entity like a flapping butterfly, the model may prioritize the rigid spatial constraint over complex non-rigid animations. This can suppress characteristic motions like wing flapping, leading to unnatural results.

\noindent\textbf{Intricate Non-Rigid Motion.} Even with high-quality multi-view priors, extreme deformation such as intricate ballet motion remains challenging because current video backbones do not reliably model complex articulated kinetics.

\subsection{Future Works}

\begin{figure*}[t]
    \centering
    \begin{subfigure}[t]{0.48\linewidth}
        \centering
        \includegraphics[width=\linewidth]{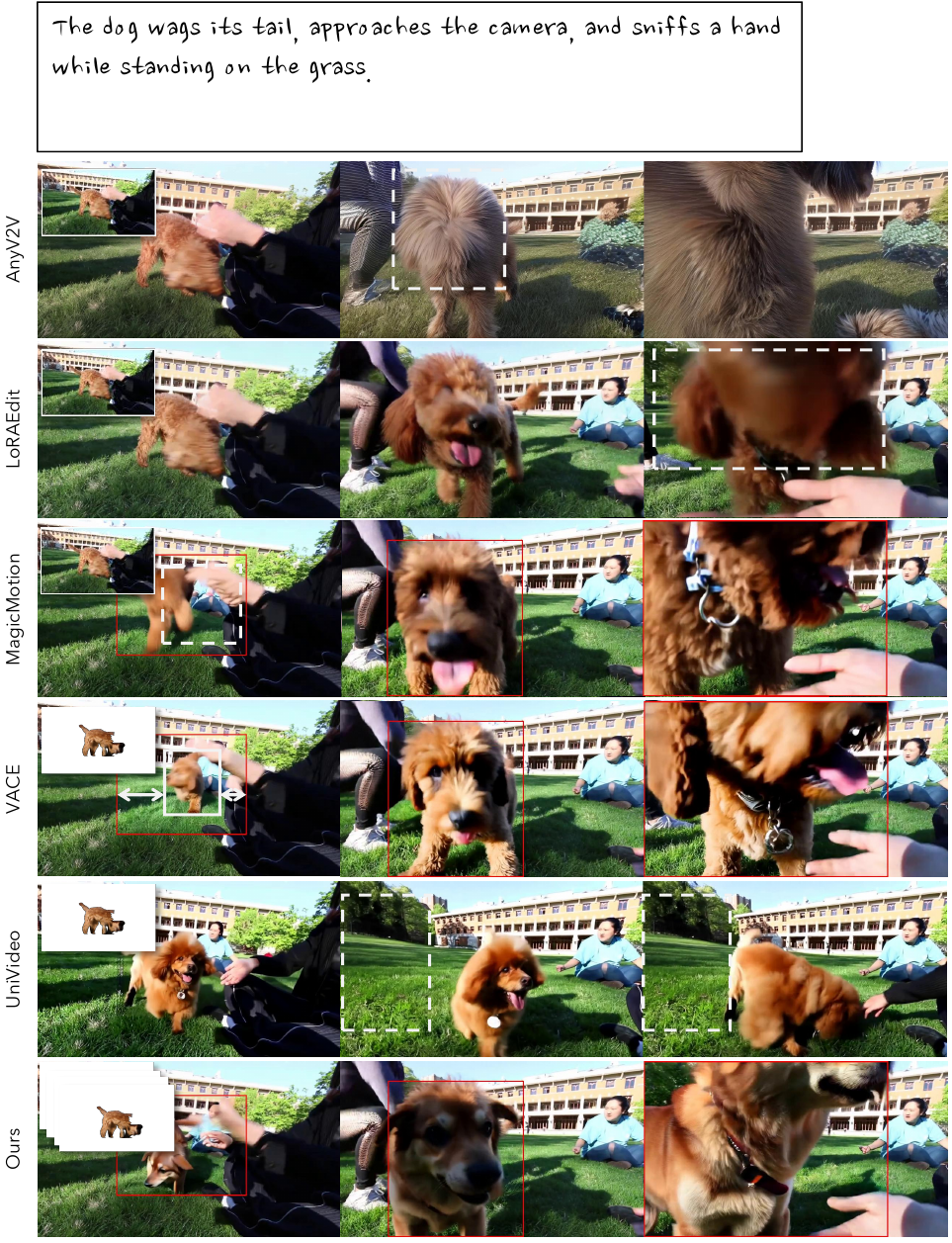}
        \caption{AnyV2V suffers from severe structural artifacts, while LoRA-Edit and MagicMotion produce blurred object-to-background boundaries. VACE fails to strictly follow the spatial control signals, and UniVideo unintentionally alters the background content. In contrast, our method faithfully adheres to the control signals, generating a clear and dynamic puppy that is seamlessly integrated into the scene.}
        \label{subfig:more_dog}
    \end{subfigure}\hfill
    \begin{subfigure}[t]{0.48\linewidth}
        \centering
        \includegraphics[width=\linewidth]{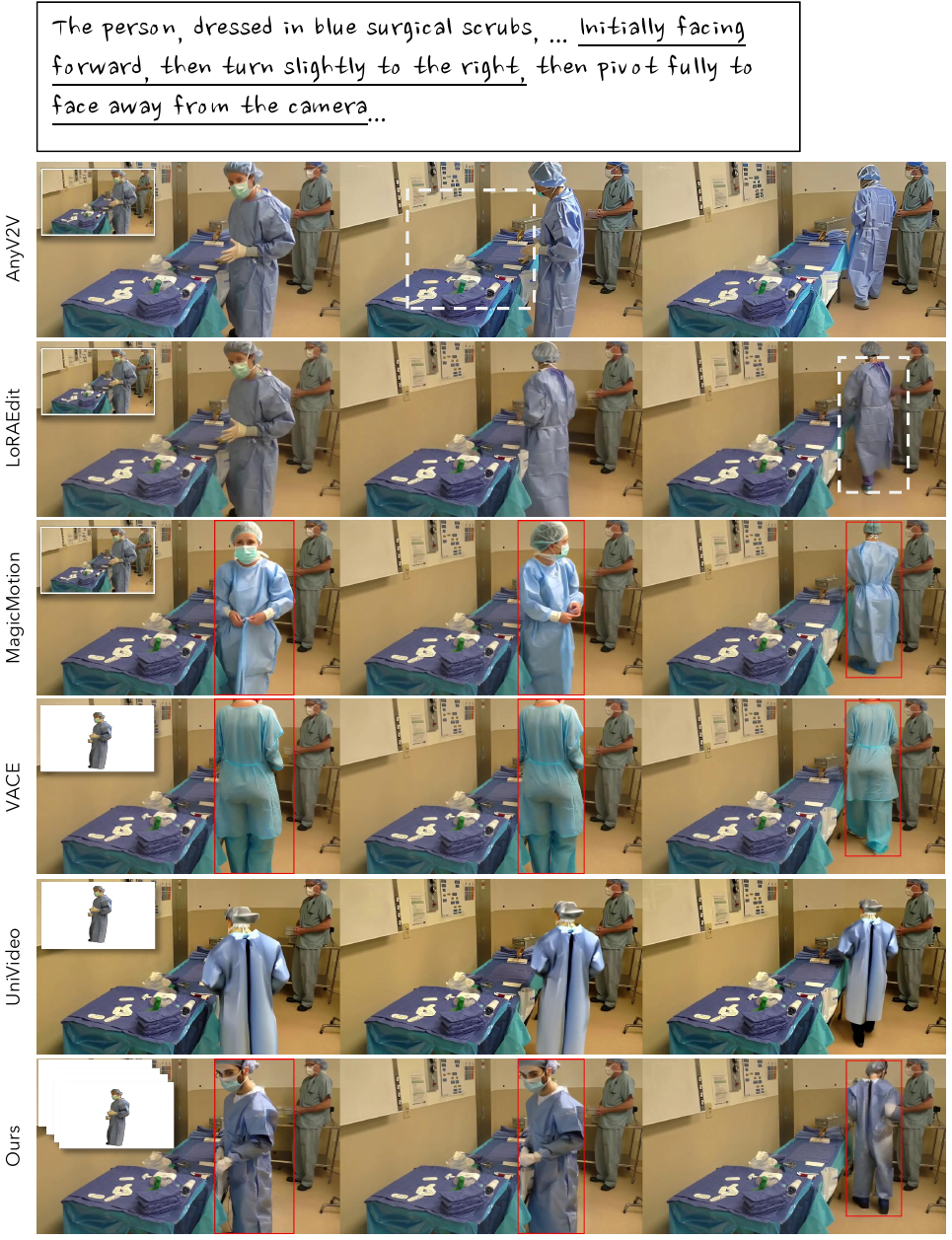}
        \caption{AnyV2V produces over-saturated textures, LoRA-Edit exhibits boundary bleeding, and VACE and UniVideo do not express the requested turning motion. Our result maintains sharper textures while synthesizing the turn.}
        \label{subfig:more_person}
    \end{subfigure}
    \caption{Additional qualitative comparisons.}
    \label{fig:additional_qualitative}
\end{figure*}

To further enhance the realism and versatility of controllable video object insertion, except for resolving the above limitations, we identify several promising directions for future research:

\noindent\textbf{Explicit Lighting and Material Harmonization.} While our framework achieves remarkable geometric consistency, achieving perfect photometric harmony between the inserted asset and the dynamic background remains a challenge. Future work will explore integrating explicit relighting modules and environment-aware material estimation to ensure that the shading, reflections, and highlights of the inserted object adaptively respond to the ambient illumination of the source video.

\noindent\textbf{Long-Range Temporal Memory.} To ensure consistency in ultra-long video sequences, we plan to investigate long-range temporal attention mechanisms or memory-augmented architectures. This will mitigate the cumulative errors in optical flow and prevent the gradual drifting of object identity during extended durations.


\balance

\end{document}